\documentclass[journal]{IEEEtran}
\ifdefined\pdfminorversion\pdfminorversion=7\fi

\usepackage{pifont}
\newcommand{\cmark}{\ding{51}}
\newcommand{\xmark}{\ding{55}}
\usepackage{amsmath,amssymb}
\newenvironment{proof}{\begin{IEEEproof}}{\end{IEEEproof}}
\usepackage{booktabs}
\usepackage{multirow}
\usepackage{tabularx}
\usepackage{array}
\usepackage{graphicx}
\usepackage{xcolor}
\usepackage{colortbl}
\usepackage{enumitem}
\usepackage{algorithm}
\usepackage[noend]{algpseudocode}
\usepackage{float}\usepackage{placeins}
\usepackage{microtype}
\usepackage{url}
\usepackage{cite}
\usepackage{xurl}
\usepackage{hyperref}
\providecommand{\citep}{\cite}
\providecommand{\Description}[1]{}
\usepackage{subfigure}

\newtheorem{ailassumption}{Assumption}[section]
\newtheorem{ailproposition}{Proposition}[section]

\definecolor{agentblue}{HTML}{183A61}
\definecolor{agentteal}{HTML}{238C83}
\definecolor{agentgold}{HTML}{D6A64B}
\definecolor{agentcoral}{HTML}{D9684B}
\definecolor{draftred}{HTML}{B94448}
\definecolor{draftbg}{HTML}{FFF4F2}
\newcommand{\code}[1]{\texttt{#1}}

\newcolumntype{Y}{>{\raggedright\arraybackslash}X}
\newcolumntype{C}{>{\centering\arraybackslash}X}
\definecolor{scoregap}{HTML}{D8A452}
\newcommand{\scorepair}[3]{\cellcolor{scoregap!#3}\shortstack[r]{\textcolor{black!60}{\normalfont\fontsize{6}{6.4}\selectfont #1}\\[-1pt]{\normalfont\fontsize{8}{8.6}\selectfont #2}}}
\title{To Optimize or Not to Optimize: When Input Optimization Helps and Hurts LLM Agents}

\author{Chuang~Zhao and Xiaomeng~Li~\IEEEmembership{Senior Member, IEEE}%
\thanks{Chuang Zhao and Xiaomeng Li are with the Department of Electronic and Computer Engineering, The Hong Kong University of Science and Technology, Hong Kong SAR, China (e-mail: czhaobo@connect.ust.hk; eexmli@ust.hk). Xiaomeng Li is the corresponding author.}}
\begin{document}
\raggedbottom 

\maketitle

\begin{abstract}
LLM agents consume heterogeneous requests, retrieved evidence, and interaction histories whose usefulness depends on how they are prepared for the task. Comparing alternative input interventions therefore requires accounting for both the task state and the execution setting. We introduce \textsc{AgentInputLab}, an open and extensible toolkit that turns input optimization from a collection of isolated techniques into a controlled decision problem: \emph{which eligible intervention, if any, should be applied to the current agent state?} Within this toolkit, we organize 23 representative algorithms into a five-family taxonomy. Our empirical study evaluates these operations across three interactive arenas on multiple local and hosted LLM backbones. 
We evaluate each intervention against a standard unoptimized baseline for its arena and backbone, as well as conducting state-matched
comparisons for representative interventions.
The central finding is that input optimization is conditional rather than uniformly beneficial: leading choices shift across arenas and execution stacks, gains vary with the observed input state, and stronger intervention intensity can reverse local gains. These setting-dependent patterns hold consistently across both Official and Valid scoring. Seven selectors built on our decision framework show how observable input signals and selective non-intervention shape routing performance relative to matched random and fixed-policy baselines. By coupling shared state and intervention interfaces with common-support paired comparisons, {\textsc{AgentInputLab}} provides reusable infrastructure for task-aware input preparation and evidence for selecting operations by their downstream utility.
\end{abstract}


\section{Introduction}
Large language model (LLM) agents extend text generation into sustained interaction: they search, call tools, update environments, maintain memory, and coordinate multi-step workflows \citep{liu2024agentbench,zhou2024webarena,li2026dataagents}. Before taking an action, an agent observes an \emph{agent-visible state} comprising the request, retrieved evidence, and interaction history. In practice, this state can suffer from poor retrieval vocabulary, obscured multi-step dependencies, missing evidence or decision variables, or an overloaded context budget. Such states can be repaired before the next action through
request rewriting~\citep{chan2024rqrag,bigdeli2026querygym},
planning~\citep{li2026globalplanning},
evidence retrieval~\citep{niu2026indexing},
clarification~\citep{aliannejadi2021clariq},
or context curation~\citep{jiang2024longllmlingua,xu2024recomp}. We
refer to these operations as \emph{input intervention}: modifying the
information or structure presented to the agent within declared task and
permission boundaries. Because the goal of this task-aware preparation is to improve task execution, we evaluate this task-aware input preparation through
downstream task outcomes. A general research toolkit should therefore decouple intervention logic from models, serving backends, retrieval stacks, tools, and dataset interfaces.

\begin{figure}[t]
\centering
\subfigure[Leaders across settings.
    \label{fig:motivation-leaders}]{%
    \includegraphics[width=0.32\columnwidth]{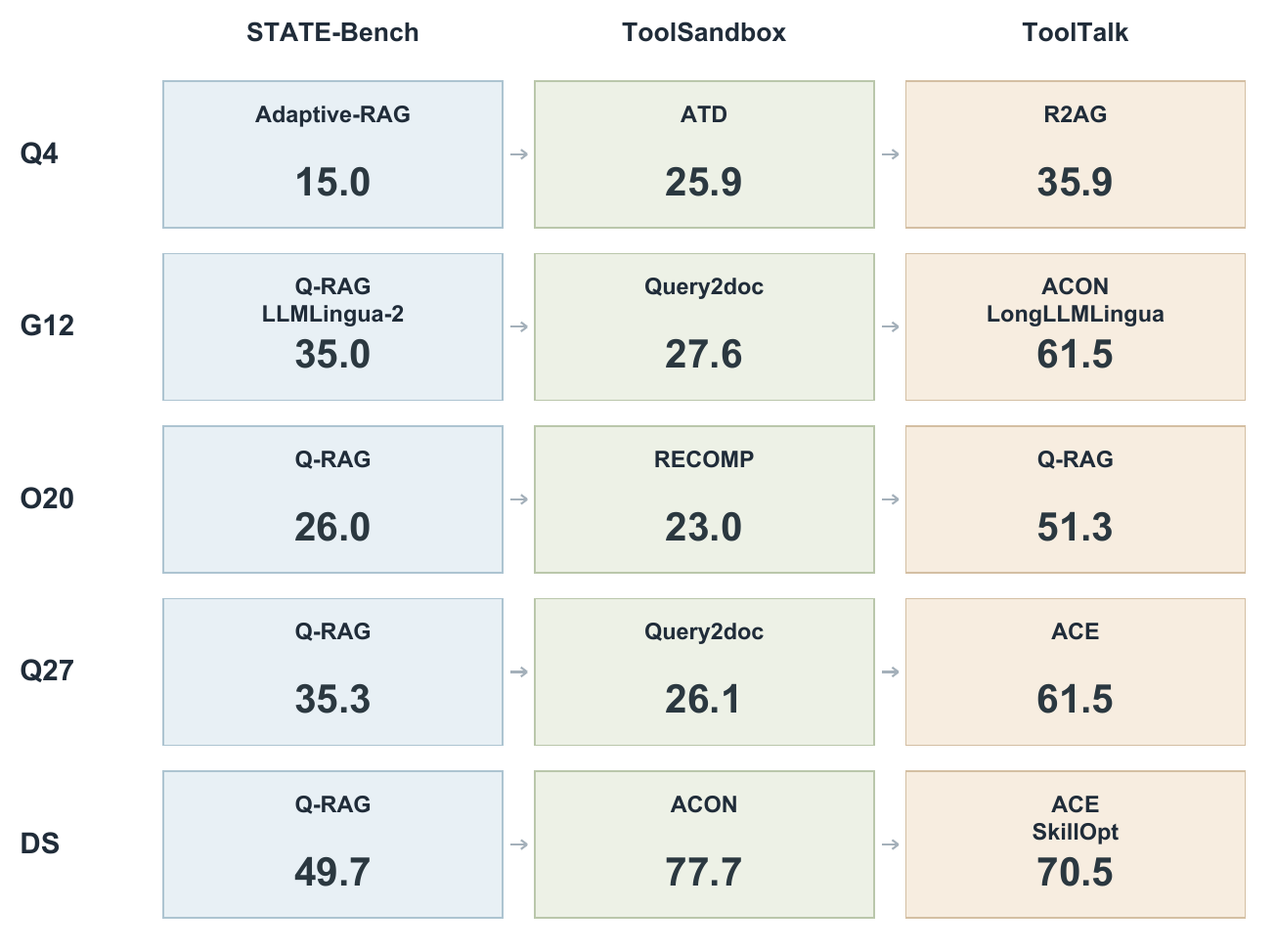}}%
\hfill
\subfigure[Task-level selection headroom.
    \label{fig:motivation-headroom}]{%
    \includegraphics[width=0.32\columnwidth]{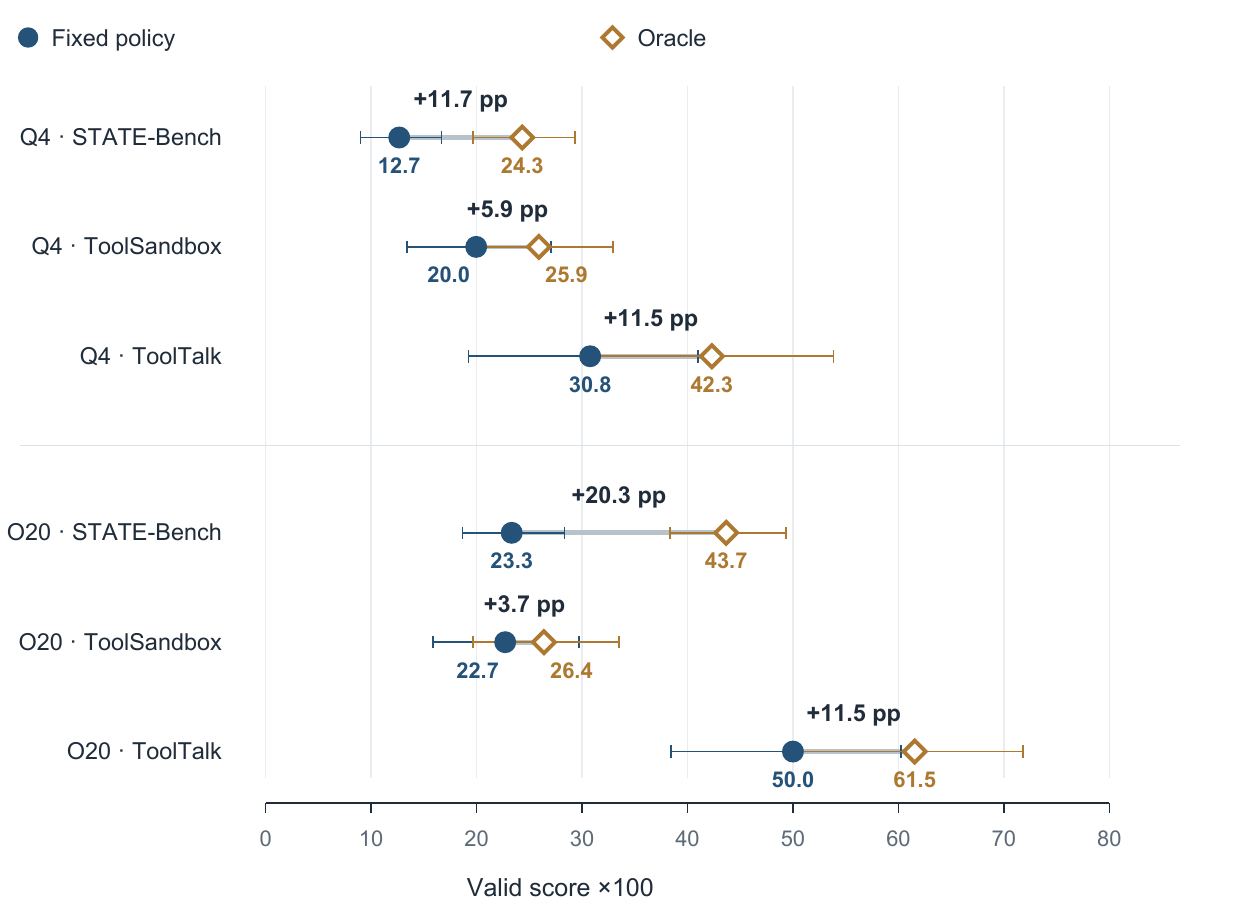}}%
\hfill
\subfigure[Task-level outcome changes.
    \label{fig:motivation-pairs}]{%
    \includegraphics[width=0.32\columnwidth]{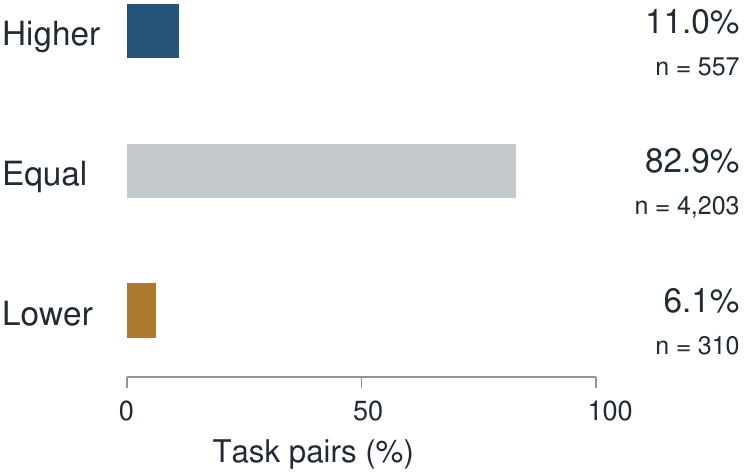}}
\caption{\textbf{Setting-dependent leaders and task-level outcomes.} (a) Top-performing pipelines (highest Valid scores, ties retained) among 23 methods and the unoptimized baseline (\code{OPT0}) across five backbones. (b) Task-wise Oracle versus a development-selected fixed action on Qwen3-4B (Q4) and GPT-OSS-20B (O20), where labels show score gaps and whiskers denote pointwise 95\% confidence intervals. (c) Valid-score gains, ties, and losses over paired \code{OPT0} across 5,070 method--task pairs. Panels (b) and (c) evaluate one representative per family (Section~\ref{sec:exp:setup}).}
\label{fig:motivation}
\end{figure}

The literature offers many effective repairs across distinct intervention roles.  Query2doc
\citep{wang2023query2doc} and Rewrite--Retrieve--Read
\citep{ma2023queryrewriting} reshape retrieval requests. Decomposed Prompting
\citep{khot2023decomposed} and LLMCompiler \citep{kim2024llmcompiler} expose
task structure. Adaptive-RAG \citep{jeong2024adaptiverag} and IRCoT
\citep{trivedi2023ircot} acquire evidence adaptively. Likewise, Ask or Assume? \citep{edwards2026askassume} resolves consequential ambiguity, while LongLLMLingua \citep{jiang2024longllmlingua} curates long context. Although these improvements are compelling within their native setups, they are not directly comparable across studies. Individual papers vary the retriever, the reader or call budget, or the user simulator and success criterion. Consequently, reported gains leave open the practical decision an agent developer faces: under matched task states, dependencies, declared resource profiles, and scorers, which intervention pipeline helps, which hurts, and when should the system leave the input untouched?

As illustrated in Fig.~\ref{fig:motivation}(a), (b) and Table~\ref{tab:methodresults}, this decision problem is governed by setting-dependent utility. Not only does the best-performing pipeline shift across arenas and backbones, but the task-wise Oracle also reveals substantial variation in
recorded outcomes beyond a development-selected fixed policy. At the task level, most paired outcomes remain unchanged, whereas aggregate averages often mask offsetting gains and losses across individual instances, as depicted in Fig.~\ref{fig:motivation}(c). A unified evaluation bridges this gap by linking outcome differences back to the initial state, intervention execution, and downstream task result, thereby pinpointing where input preparation actually pays off. Yet establishing such a credible comparison remains difficult for four reasons that existing benchmarks address only in part. \textbf{Scope heterogeneity:} rewriting a query, adding evidence, asking a user, and deleting context modify different state fields and are not applicable in every state. QueryGym \citep{bigdeli2026querygym} is an important step toward reproducible query-reformulation research: it standardizes nine reformulators across datasets, retrievers, and LLMs, and a follow-up study demonstrates substantial sensitivity to these choices \citep{bigdeli2026reproducibility}. Its controlled scope, however, remains a retrieval query rather than the broader interactive state altered across intervention families.   \textbf{Dependency confounding:} even within retrieval, chunking, indexing, the retriever, top-$k$, and resource budgets can reverse a conclusion, making it difficult to attribute gains to the intervention alone. Our retriever results in Section~\ref{sec:rq4} show that the same query-expansion method yields markedly different gains across retrieval stacks, while Fig.~\ref{fig:budget-recovered} reveals distinct budget responses across methods and arenas.   \textbf{State and execution consistency:} general agent benchmarks such as AgentBoard \citep{ma2024agentboard} and ToolSandbox \citep{lu2025toolsandbox} evaluate end-to-end trajectories, yet they do not isolate the pre-action intervention as a paired treatment. Without restoring both runs from a shared checkpoint, environmental drift can confound the measured difference. The gaps between Official and Valid scores in Table~\ref{tab:methodresults} also separate task success from intervention completion. For example, an InfoCQR run failed to rewrite the request and returned it unchanged, yet the downstream agent still received full task credit, as detailed in Supplement~\ref{app:component}.   \textbf{Conditional utility:} existing routing methods select retrieval strategies or reasoning paradigms~\citep{jeong2024adaptiverag,zhou2026selectthensolve}, but they do not address choosing among heterogeneous, sometimes ineligible interventions without leaking post-decision outcomes. They also leave open when to abstain via \code{OPT0}, an identity action that leaves the input unchanged. Fair comparison therefore demands more than a shared leaderboard: it requires a common state contract, controlled dependencies, paired execution, and an explicit decision boundary.

\textsc{AgentInputLab} addresses these four challenges through a coordinated design. To handle \emph{scope heterogeneity}, it organizes 23 representative algorithms into five intervention families: \textit{reformulation}, \textit{decomposition}, \textit{evidence acquisition}, \textit{clarification}, and \textit{context curation}, as summarized in Table~\ref{tab:methods}. Each intervention returns a typed patch recording its input changes and runs only when the required information and interfaces are available. We compare interventions on the same task groups. \emph{Dependency confounding} is addressed through native, controlled, and factorial evaluation profiles, introduced in Section~\ref{sec:profiles}, where controlled comparisons match models, retrieval stacks, tools, declared resource profiles, and scorers, while factorial profiles vary one dependency at a time. To ensure \emph{state and execution consistency}, paired runs independently restore from the same checkpoint and link task scores to intervention-completion and information-access traces, as described in Equation.~\ref{eq:estimand}. Finally, \emph{conditional utility} is modeled through seven selectors that rely only on pre-action features, enforce intervention eligibility, and retain \code{OPT0} as a valid action.   Beyond this core setup, the toolkit also reuses its interfaces across task-specific retrieval, document-QA, and summarization settings in Section~\ref{sec:rq1} and Supplement~\ref{app:auxiliary-four}, whose setting-dependent gains align with our main findings. Among the benchmarks introduced in Section~\ref{sec:benchmarks}, STATE-Bench \citep{microsoft2026statebench}, ToolSandbox \citep{lu2025toolsandbox}, and ToolTalk~\citep{farn2023tooltalk} offer task instances where all five families can operate, enabling cross-family comparison on common task groups.  

Across our experiments, rankings remain task-dependent under both Official and Valid scoring (Section~\ref{sec:rq4}), while the fixed-task repeats in Section~\ref{sec:rq2} distinguish recurring gain directions from run-to-run sensitivity. As Fig.~\ref{fig:mechanisms} illustrates, higher intervention intensity can reverse an improvement, showing that simply dialing up intervention is no substitute for picking the right method and operating point. Building on these patterns, the routing analysis in Fig.~\ref{fig:selector} ties task-level utility to observable input signals, positioning non-intervention as a core routing choice rather than a fallback exception.   
Our contributions are fourfold:
\begin{itemize}[leftmargin=*,nosep]

\item {
We develop a unified evaluation protocol whose comparison unit
links an agent-visible state, an eligible intervention pipeline,
a declared execution profile, and a downstream task outcome.
Implemented in \textsc{AgentInputLab} with 23 method adapters,
the protocol supports both aggregate pipeline comparisons and task-level
comparisons against state-matched \code{OPT0}. It links task scores to
intervention completion, resource usage, and execution traces.
}

\item {
We establish an operational taxonomy of five input-intervention
families, specifying their target fields, capability requirements,
and comparison boundaries. Typed patches and eligibility checks
make interface compatibility explicit, allowing the applicability
of an operation to be distinguished from its measured
downstream benefit.
}

\item {
We formulate conditional intervention as a selection problem
and use seven development-calibrated policies as diagnostic
baselines. Task-group-disjoint splits, rate-matched random
controls, development-selected fixed actions, and feature-block
ablations support evaluation of whether observable input
signals justify varying interventions across tasks.
}

\item {
Our empirical study identifies three practical distinctions:
downstream task success can coexist with incomplete intervention
execution; increasing transformation strength under a fixed
total budget can reduce utility; and retrospective selection
headroom can coexist with fixed policies that match or
outperform learned selectors. These findings inform integration
checks, intervention calibration, and the choice between fixed
and input-dependent policies.
}

\end{itemize}
The \href{https://data-designer.github.io/AgentInputLab/}
{project website} provides tutorials, while the
\href{https://github.com/Data-Designer/AgentInputLab-toolkit}
{public repository} and
\href{https://pypi.org/project/agentinputlab/}
{PyPI package} provide the toolkit source and installation
entry points.

\section{Related Work}
\textsc{AgentInputLab} sits at the intersection of agent input intervention and unified evaluation.

\begin{table*}[ht]
\setlength{\abovecaptionskip}{-0.05cm}
\setlength{\belowcaptionskip}{-0.1cm}
\caption{The 23 methods evaluated in this study. ``Changed object'' determines
the primary family, and \code{OPT0} denotes the unchanged-input ReAct reference. Native
method logic and configurations remain traceable to the cited work.}
\label{tab:methods}
\small
\setlength{\tabcolsep}{4.0pt}
\begin{tabularx}{\textwidth}{llclY}
\toprule
Family & Method & Year & Primary changed object & Core mechanism \\
\midrule
\multirow{4}{*}{Reformulate}
& Query2doc \citep{wang2023query2doc} & 2023 & Query & Append a generated pseudo-document. \\
& Rewrite-Retrieve-Read \citep{ma2023queryrewriting} & 2023 & Query & Rewrite before retrieval. \\
& InfoCQR \citep{ye2023infocqr} & 2023 & Conversational query & Make a dialogue query standalone. \\
& ReFormeR \citep{bigdeli2026reformer} & 2026 & Query & Apply learned rewrite patterns. \\

\midrule
\multirow{5}{*}{Decompose}
& Plan-and-Solve \citep{wang2023planandsolve} & 2023 & Plan & Plan before solving. \\
& Decomposed Prompting \citep{khot2023decomposed} & 2023 & Subtasks & Delegate subtasks to handlers. \\
& Self-Ask \citep{press2023selfask} & 2023 & Follow-up questions & Alternate follow-up queries and search. \\
& LLMCompiler \citep{kim2024llmcompiler} & 2024 & Execution graph & Compile a graph of function calls. \\
& Plan-and-Budget \citep{lin2026planbudget} & 2026 & Plan and budget & Allocate computation across steps. \\
\midrule
\multirow{4}{*}{Acquire}

& IRCoT \citep{trivedi2023ircot} & 2023 & Evidence chain & Alternate reasoning and retrieval. \\
& Adaptive-RAG \citep{jeong2024adaptiverag} & 2024 & Retrieval route & Choose retrieval depth by complexity. \\
& $R^2$AG \citep{ye2024r2ag} & 2024 & Retrieval representation & Inject retrieval features. \\
& Q-RAG \citep{sorokin2026qrag} & 2026 & Long-context evidence & Select evidence using learned values. \\
\midrule
\multirow{3}{*}{Clarify}

& Active Task Disambiguation \citep{kobalczyk2025active} & 2025 & Clarification question & Choose questions by information gain. \\
& Ask or Assume? \citep{edwards2026askassume} & 2026 & Clarification state & Ask when an assumption is risky. \\
& UDCS \citep{matsnev2026udcs} & 2026 & Clarification state & Ask targeted uncertainty questions. \\
\midrule
\multirow{7}{*}{Curate}
& LongLLMLingua \citep{jiang2024longllmlingua} & 2024 & Long context & Compress context around the query. \\
& LLMLingua-2 \citep{pan2024llmlingua2} & 2024 & Prompt tokens & Classify tokens for retention. \\
& RECOMP \citep{xu2024recomp} & 2024 & Retrieved context & Select and compress retrieved evidence. \\
& ACON \citep{kang2026acon} & 2026 & Agent context & Learn long-task context compression. \\
& ACE \citep{zhang2026ace} & 2026 & Context policy & Update context from feedback. \\
& MCE \citep{ye2026mce} & 2026 & Context skills & Refine context-engineering skills. \\
& SkillOpt \citep{yang2026skillopt} & 2026 & Executable skill context & Optimize reusable agent skills. \\
\bottomrule
\end{tabularx}
\end{table*}

\subsection{Input Intervention for LLM Agents}
Input intervention alters the information shown to an LLM agent before its next action through request, evidence, interaction, or context updates \citep{wang2023query2doc,bigdeli2026querygym,ma2023queryrewriting}. Defined by its pre-action timing rather than a fixed mechanism, it seeks to strengthen the agent's decision state without violating user intent, permissions, or constraints.

Prior work has evolved largely within isolated, task-specific threads. Retrieval-focused approaches reshape query--evidence interfaces: Query2doc \citep{wang2023query2doc} appends pseudo-documents, Rewrite--Retrieve--Read \citep{ma2023queryrewriting} couples rewriting with retrieval, and HyDE \citep{gao2023hyde} and $R^2$AG \citep{ye2024r2ag} synthesize or inject stronger retrieval signals. Reasoning-focused approaches restructure execution via planning, subproblem delegation, or dependency graphs, exemplified by Plan-and-Solve \citep{wang2023planandsolve}, Decomposed Prompting \citep{khot2023decomposed}, Self-Ask \citep{press2023selfask}, and LLMCompiler \citep{kim2024llmcompiler}. Conversational query rewriting methods such as InfoCQR
\citep{ye2023infocqr} make dialogue-dependent requests self-contained,
while interactive clarification methods such as Ask or Assume?
\citep{edwards2026askassume} and UDCS \citep{matsnev2026udcs} seek
missing information from the user. Context-curation methods such as
LongLLMLingua \citep{jiang2024longllmlingua}, LLMLingua-2
\citep{pan2024llmlingua2}, RECOMP \citep{xu2024recomp}, and ACON
\citep{kang2026acon} manage scarce context budgets. 
Related work also examines how knowledge and context can be prepared for generation. Frameworks integrating LLMs with knowledge graphs treat external knowledge as an inference resource \citep{pan2024roadmap}, while finer-grained methods decompose outputs into atomic facts for evidence retrieval and revision \citep{yan2025atomic} or inject verified knowledge at each mathematical reasoning step \citep{liu2025knowledge}. Work on context compression \citep{xu2024recomp} and demonstration valuation \citep{zhou2025valuing} further ties input selection to downstream performance, underscoring the need to evaluate prepared inputs within the consuming task and system.

Although these studies share a pre-action intervention point, they differ in their objectives, system configurations, and evaluation settings.
To enable such comparison, \textsc{AgentInputLab} introduces an operational five-family taxonomy organized by the modified agent-visible object and required capabilities (Table~\ref{tab:methods} and Supplement~\ref{app:overview}). Heterogeneous methods are mapped to typed patches with explicit eligibility checks, while adapters expose their internal stages and declared dependency settings. 
Paired execution  in Section~\ref{sec:rq2} compares each of the five representative interventions with
\code{OPT0} from the same restored state under matched dependencies,
bringing otherwise fragmented results into a unified comparative framework.

\begin{table}[!h]
\setlength{\abovecaptionskip}{-0.05cm}
\setlength{\belowcaptionskip}{-0.1cm}
\setlength{\tabcolsep}{2.2pt}
\caption{Comparison with representative evaluation frameworks.
\cmark, \xmark, and $\triangle$ denote full, absent, and partial
support, respectively; partial plugin support denotes
module-level extension within a specific workflow.
``5F'' denotes coverage of all five input-intervention families;
R denotes reformulation only.
Paired evaluation restores each intervention--reference pair
from the same environment checkpoint.
``Select.'' denotes task-dependent method or strategy selection.}
\label{tab:relatedbench}
\scriptsize
\begin{tabularx}{\columnwidth}{@{}lXcccc@{}}
\toprule
Toolkit & Focus & Plugins & 5F & Paired & Select. \\
\midrule
HELM \citep{liang2023helm}
& LMs/scenarios
& \cmark & \xmark & \xmark & \xmark \\

AgentBench \citep{liu2024agentbench}
& Agents
& $\triangle$ & \xmark & \xmark & \xmark \\

AgentBoard \citep{ma2024agentboard}
& Trajectories
& $\triangle$ & \xmark & \xmark & \xmark \\

WebArena \citep{zhou2024webarena}
& Web agents
& $\triangle$ & \xmark & \xmark & \xmark \\

QueryGym \citep{bigdeli2026querygym}
& Reformulation
& \cmark & R & \xmark & \xmark \\

QueryRoute \citep{queryroute}
& Reformulation selection & $\triangle$ & R & \xmark & \cmark \\
Select-then-Solve \citep{zhou2026selectthensolve}
& Paradigm routing & $\triangle$ & \xmark & \xmark & \cmark \\

\textbf{\textsc{AgentInputLab}}
& \textbf{Interventions}
& \cmark & \cmark & \cmark & \cmark \\
\bottomrule
\end{tabularx}
\end{table}

\subsection{Benchmarking and Conditional Selection}
Evaluation toolkits facilitate reproducible model and agent comparisons by standardizing tasks, interfaces, metrics, and reporting \citep{ma2024agentboard,liu2024agentbench}. Such standardization is especially important for agents, whose outcomes depend on multi-step trajectories, tool calls, mutable environments, and resource constraints. Conditional intervention introduces further demands: benchmarks must provide comparable alternative actions, explicit eligibility rules, and leakage-free pre-intervention selection.
Separately, fine-grained evaluation research highlights the value of decoupling process fidelity from final answers, such as separating answer correctness from explanation quality and distinguishing evidence-selection errors from reasoning errors in logical reasoning \citep{xu2025logical}. This perspective aligns closely with our distinction between downstream task outcomes and intervention completion, which \textsc{AgentInputLab} embeds directly into its paired evaluation protocol.

Existing toolkits extend standardized evaluation from language
models to interactive agents. HELM \citep{liang2023helm}
introduced broad multi-scenario and multi-metric evaluation
for models, while AgentBench \citep{liu2024agentbench},
AgentBoard \citep{ma2024agentboard}, and WebArena
\citep{zhou2024webarena} evaluate end-to-end agents in interactive
environments. ToolSandbox \citep{lu2025toolsandbox} and BFCL
\citep{patil2025bfcl} focus on stateful tool use and function
calling, respectively. QueryGym \citep{bigdeli2026querygym}
offers a close software precedent for input optimization by
integrating nine query-reformulation methods with pluggable
retrievers, multiple retrieval datasets, and reproducible run
schemas. Its follow-up reproducibility study
\citep{bigdeli2026reproducibility} further demonstrates that
reformulator rankings shift across model families, scales,
retrievers, and datasets.
Related work also examines conditional selection, including
retrieval-strategy selection in Adaptive-RAG
\citep{jeong2024adaptiverag}, reasoning-paradigm selection in
Select-then-Solve \citep{zhou2026selectthensolve}, query-variant
selection via QPP \citep{arabzadeh2026qppvariant}, and risk-aware
abstention in MERA \citep{hedstrom2025mera}. QueryRoute
\citep{queryroute} evaluates query-reformulation selection using
frozen retrieval outcomes. 

\textsc{AgentInputLab} provides a common framework for comparing
different types of input intervention by their downstream task
performance. General agent benchmarks evaluate complete systems,
while QueryGym and QueryRoute focus on query reformulation and
Select-then-Solve on reasoning paradigms. Our toolkit covers
reformulation, decomposition, evidence acquisition, clarification,
and context curation, enabling cross-family comparisons on common
task groups. We further use selection baselines to examine whether
information available before intervention can guide the choice
among applicable methods and \code{OPT0}, which leaves the input
unchanged. This allows us to study both when different forms of
input preparation help or hurt and when task-dependent selection
adds value over a fixed strategy.
Table~\ref{tab:relatedbench} summarizes these differences.

\section{\textsc{AgentInputLab}}
\label{sec:framework}
\textsc{AgentInputLab} identifies which interventions can run
in a given task state and evaluates their downstream task
performance under specified execution settings.
The framework supports comparisons with \code{OPT0}
and intervention selection using only information available
before intervention.

\begin{figure}[t]
\centering
\includegraphics[width=\linewidth]{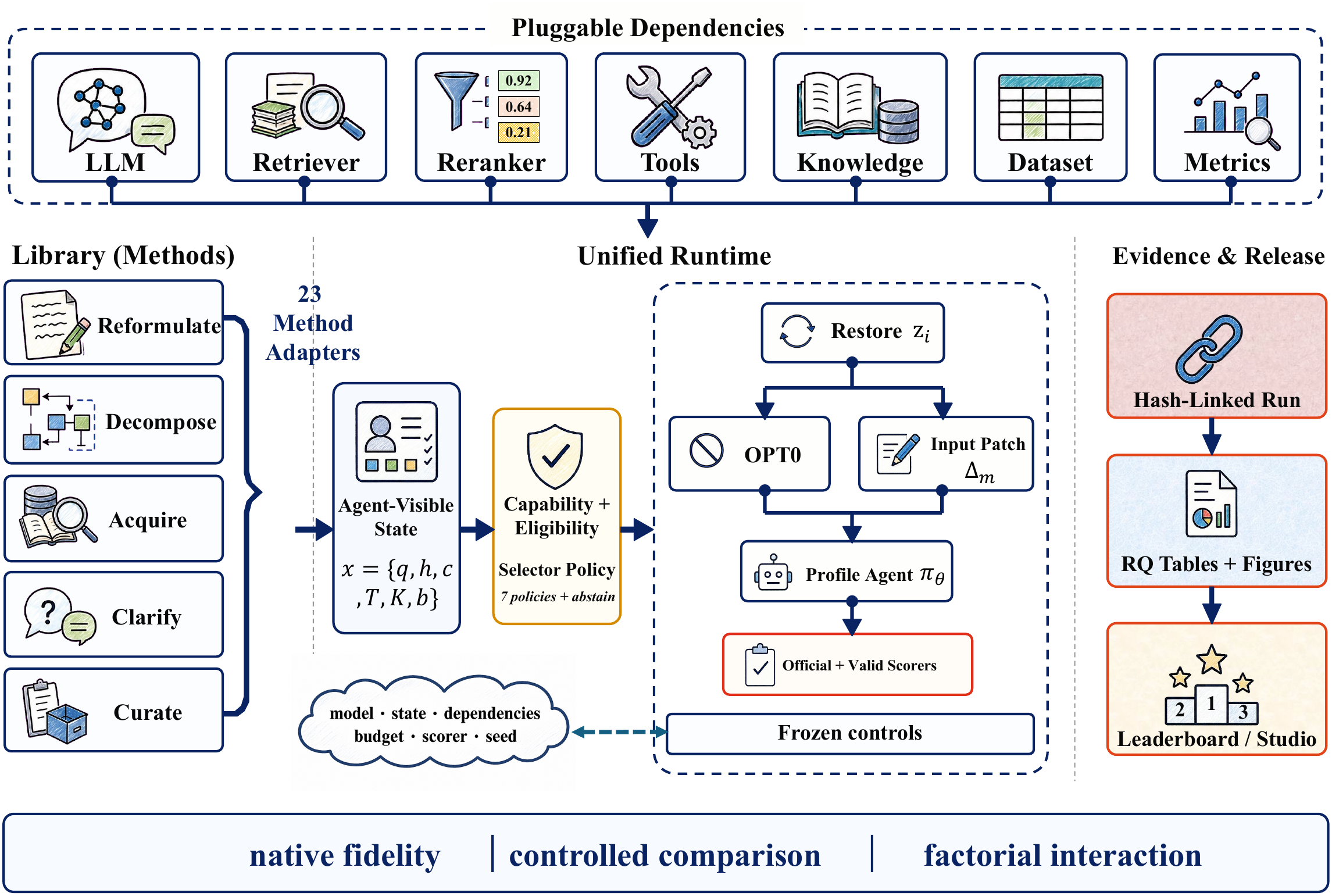}
\setlength{\abovecaptionskip}{-0.01cm}   
\setlength{\belowcaptionskip}{-0.1cm}   
\caption{\textbf{\textsc{AgentInputLab}.}
Plugins connect through typed interfaces.
The runtime checks intervention requirements, restores task
states for paired comparisons, and runs interventions and
downstream agents under specified settings.
Execution logs support evaluation, selection analysis,
and reporting.}
\label{fig:framework}
\vspace{-1em}
\end{figure}

\subsection{Problem Definition}
We index tasks by $i$ and candidate interventions by $m$. Let $x_i$ denote the
observable pre-action state, including the request, dialogue history, visible
context, available tools, permitted knowledge sources, and remaining budget.
Let $z_i$ denote the protected environment snapshot and task
specification. A comparison profile $\theta$ specifies the shared model, retrieval
stack, tools, decoding settings, and scorer, as well as each arm's resource
limits fixed within a comparison block.

For an eligible method $m$, let $I_m$ denote its intervention operator.
Applying $I_m$ to $x_i$ under profile $\theta$ yields a typed patch
$\Delta_{i,m}$,  which summarizes input changes across one or more internal steps and records modified fields, provenance, dependencies, and resource usage. The operator $\oplus$ applies this patch to the original state,
and $\tilde{x}_{i,m}$ denotes the resulting intervened state:
\begin{equation}
  \Delta_{i,m}=I_m(x_i;\theta), \qquad
  \tilde{x}_{i,m}=x_i\oplus\Delta_{i,m}, \qquad
  I_0(x_i;\theta)=\varnothing .
  \label{eq:intervention}
\end{equation}
Here, $m=0$ denotes the identity intervention, or \code{OPT0}, whose patch is
empty. Each $\Delta_{i,m}$ records the fields modified by the intervention
along with its provenance, dependencies, and resource use. The eligible action set $\mathcal{E}(x_i)$ contains methods whose capability requirements are satisfied by state $x_i$, with \code{OPT0} always included. 
Interventions may modify only the observable state and cannot directly mutate
the protected snapshot $z_i$.
Let $\pi_{\theta,m}$ denote the downstream policy associated with
method $m$ under comparison profile $\theta$. For each
method--reference pair, both arms independently restore the same
protected snapshot, denoted by $z_i$ within that pair. The
intervention arm receives $\tilde{x}_{i,m}$ and produces
$\tau_{i,m}=\pi_{\theta,m}(\tilde{x}_{i,m},z_i)$; the unchanged-input
arm receives $x_i$ and uses the ReAct policy $\pi_{\theta,0}$ to
produce $\tau_{i,0}$. When initial environments differ across
methods, the snapshot and reference are indexed within each pair.
We instantiate the unchanged-input action, \code{OPT0}, with ReAct\citep{yao2023react}: its
tool-feedback loop provides an executable reference without an input-optimization
stage. We use \code{OPT0} for this reference below.
For instance, InfoCQR inspects the request and dialogue history and returns a request-field patch, after which the downstream agent runs from the restored snapshot and the native scorer grades the resulting trajectory.

Let $Y^O_{i,m}$ denote the official task score.
We set $G_{i,m}=1$ if an accepted score is available, the method reports no algorithmic failure, and the recorded
checks detect no leakage of privileged information; otherwise, $G_{i,m}=0$.
Documented failed runs without an accepted score are included in aggregate scores as zeros.
We term $Y^T_{i,m}$ the \emph{Valid score} and
define it alongside the paired system difference $\delta_m$ as
\begin{equation}
  Y^T_{i,m}=Y^O_{i,m}G_{i,m}, \qquad
  \delta_m=\mathbb{E}\!\left[Y^T_{i,m}-Y^T_{i,0}
  \mid m\in\mathcal{E}(x_i)\right].
  \label{eq:estimand}
\end{equation}
As an example, a failed rewrite can return the original request and still yield a successful tool action. Official scoring awards credit for that action, while $G_{i,m}=0$ zeros out the Valid score owing to incomplete intervention execution. In the paired analysis, the difference $\delta_m$ averages valid-score gains or losses over paired states where $m \in \mathcal{E}(x_i)$, comparing the intervention pipeline with \code{OPT0} across their respective downstream policies and resource profiles. Every paired execution further records the official score, execution flags, latency, token counts, model and tool calls, and estimated cost, following the scoring and resource rules in Section~\ref{sec:exp:setup}.

\subsection{Taxonomy of Input Interventions}
We introduce an operational taxonomy that classifies each method by its primary
semantic operation on the agent-visible state and the capabilities it requires.
This distinguishes interventions that modify similar state fields but serve
different roles. Hybrid methods may keep secondary mechanism tags, while their primary family determines the typed patch, eligibility conditions, and method-specific input-demand criterion. \code{OPT0} remains a benchmark action rather than an intervention family.
We represent the agent-visible state as
$x=(q,h,c,\mathcal{T},\mathcal{K},b)$, denoting the current request, interaction
history, visible context, available tools, permitted knowledge sources, and
remaining budget, respectively. For each family $F$, $g_F$ denotes its
transformation operator. We use $x[s\leftarrow v]$ to replace field $s$ with
$v$, and $\oplus$ for typed append or composition. The five intervention
families are formalized as follows:
\begin{alignat}{2}
\textsc{Reformulate:}\;
  &q' = g_R(q,h), &&\; x^R=x[q\leftarrow q']; \\
\textsc{Decompose:}\;
  &p = g_D(q,h,\mathcal{T}), &&\; x^D=x[c\leftarrow c\oplus p]; \\
\textsc{Acquire:}\;
  &r = g_A(q,h,\mathcal{K}), &&\; x^A=x[c\leftarrow c\oplus r]; \\
\textsc{Clarify:}\;
  &(u,a)=g_C(q,h), &&\; x^C=x[h\leftarrow h\oplus(u,a)]; \\
\textsc{Curate:}\;
  &c' = g_U(c;q,h), &&\; x^U=x[c\leftarrow c'].
\end{alignat}
Above, $q'$, $p$, $r$, $(u,a)$, and $c'$ denote the reformulated request, task
structure, acquired evidence, clarification exchange, and curated context,
respectively. Reformulation targets the request. Decomposition primarily adds
task structure, whereas acquisition primarily augments evidence from declared
knowledge sources. Clarification may add an interaction turn, and curation may
remove, compress, or reorder context.
These boundaries determine intervention eligibility in each benchmark state,  with Table~\ref{tab:methods} listing the inventory and Supplement~\ref{app:interface} detailing the interfaces.

\subsection{System and Evaluation Protocol}
Fig.~\ref{fig:framework} summarizes the implementation. The toolkit provides typed interfaces for methods, datasets,
model backends, retrievers, metrics, and selectors.
Adapters convert the shared task state into each method's
input format and return patches for the specified state fields.
Models, retrievers, and other dependencies are configured
separately from method implementations, allowing them to
be held fixed in controlled comparisons.

\noindent\textbf{Interchangeable dependencies.}
\textsc{AgentInputLab} separates intervention logic from replaceable execution
dependencies. Its language-model interface supports local Transformers
inference \citep{wolf2020transformers}, vLLM serving
\citep{kwon2023vllm}, other OpenAI-compatible servers, and native provider
APIs through a shared request and response schema, so compatible backends
can be selected through configuration. Retrieval-dependent methods use
explicit embedding, sparse and dense retrieval, reranking, and vector-search
interfaces, with matched profiles for BM25
\citep{robertson2009probabilistic}, BGE \citep{xiao2024cpack}, E5
\citep{wang2022textembeddings}, and Qwen3-Embedding
\citep{zhang2025qwen3embedding}. Controlled comparisons fix the corpus,
chunking, index, retriever, reranker, top-$k$, and reader. Benchmark adapters
separately declare available tools, permissions, reset operations, knowledge
snapshots, and scorers, ensuring that methods request only supported
capabilities in Section~\ref{sec:exp:setup}. Optional upstream repositories run in fixed worker or Docker
environments \citep{merkel2014docker}. Each execution records model, data,
retriever, prompt, budget, and code identities alongside token and call
usage, latency, memory, failures, and the resulting patch. Fixed run records and hash-linked logs then connect every run to its tables, figures, and
leaderboard entries, supporting one-command execution and verification
of paired comparisons. For usability, the installable \href{https://pypi.org/project/agentinputlab/}{Python package} connects dependency setup, configuration, execution, and native scoring for supported combinations, while a web interface exposes run controls and preflight checks catch missing inputs or unsupported setups before execution begins.

\noindent\textbf{Fairness by design.}\label{sec:profiles}
Three evaluation profiles serve distinct purposes in our framework. The \emph{native} profile retains each method's released dependencies for upstream-conformance checks. The \emph{controlled} profile standardizes the dependency stack for matched cross-method comparisons. The \emph{factorial} profile varies a single dependency while keeping the remaining ones fixed to isolate its sensitivity.

Experiment-specific profile choices appear in Section~\ref{sec:exp:setup}, and Supplement~\ref{app:fair} formalizes the execution boundary along with the settings fixed inside controlled comparison blocks. Component-level verification across Q4 and O20 on all three arenas is summarized in Table~\ref{tab:comparison-evidence}, covering executed rewrites, planning triggers, document selection, clarification transitions, and frozen skill content, with detailed traces and fixed-task repeats in Supplement~\ref{app:component}.
Fig.~\ref{fig:protocol} in Supplement~\ref{app:overview} connects the capability interfaces to our comparison blocks, each binding a task checkpoint, visible state, model configuration, intervention, resource budget, and scorer.

\subsection{Dataset and Evaluation Interfaces}
\label{sec:benchmarks}

\textsc{AgentInputLab} uses dataset plugins to normalize static examples and restorable interactive episodes into a shared task contract, exposing visible fields, capabilities, provenance, and official scorers without altering method adapters. Table~\ref{tab:mainbench} in Supplement~\ref{app:overview} summarizes the supported interfaces across interactive agents, retrieval, multi-hop and long-document QA, and summarization, distinguishing the seven evaluated settings from other registered benchmarks. Task selection balances computational cost against comparable support across the interventions under study. Our primary study focuses on STATE-Bench \citep{microsoft2026statebench}, ToolSandbox \citep{lu2025toolsandbox}, and ToolTalk~\citep{farn2023tooltalk}, where prespecified instances support all five intervention families. Cross-family comparisons are restricted to these common-support instances, with eligibility and execution status logged explicitly. By contrast, the remaining interfaces support subsets of the intervention families and serve separate task-specific evaluations, including the retrieval, document question answering, and summarization studies reported in Supplement~\ref{app:auxiliary-four}.

Evaluation combines official task metrics, where applicable
\citep{thakur2021beir,su2025bright,trivedi2022musique,ho2020twowiki,yang2018hotpotqa,dasigi2021qasper},
with automated process checks.
We report the official task score $Y^O$ and the Valid score $Y^T$.
Exact matching of entities, numbers, and constraint spans
measures lexical retention of protected text.
The process checks also record algorithm failure types,
adapter stops, and leakage-check results.
Additional measurements include runtime failures, resource
usage, latency, memory use, and estimated cost.
Metric plugins define how to aggregate results, handle missing
values, and record the source of each result.
Interface details are summarized in Supplement~\ref{app:overview}.

\section{Experiments}
\label{sec:experiments}
In our empirical experiments, we compare aggregate pipeline scores, use paired runs to examine
task-level gains and losses, and evaluate conditional routing.

\subsection{Research Questions and Study Logic}
Our empirical study investigates five research questions:
\begin{description}[leftmargin=!,labelwidth=3.2em,nosep]
\item[RQ1] How do the 23 evaluated methods compare within each arena--backbone setting under the declared execution profiles?
\item[RQ2] Starting from the same task state, when does a pipeline improve over \code{OPT0}, and does it yield larger gains when an input exhibits its targeted need (\emph{Matched}) rather than on other inputs (\emph{Off-target})?
\item[RQ3] How do task scores change when a method plans more steps, retrieves more evidence, or compresses more aggressively, and how do token and call budgets shape these outcomes?
\item[RQ4] How do method rankings vary across backbones and arenas, and how sensitive are outcomes to retrieval and resource configurations?
\item[RQ5] Does selecting a pipeline from the input outperform random selection at the same intervention rate, and which pre-action signals guide that choice across arenas and backbones?
\end{description}

These questions follow a progressive study logic: RQ1 establishes setting-specific pipeline performance, while RQ2 uses restored-state pairs to uncover task-level gains and losses masked by aggregate averages. RQ3 and RQ4 then examine how intervention intensity and execution dependencies alter those comparisons. Building on these findings, RQ5 tests whether pre-action input signals support conditional selection beyond random baselines under matched intervention rates.

\subsection{Experimental Setup}\label{sec:exp:setup}

\noindent\textbf{Datasets, backbones, and deployment.}
We evaluate 23 intervention pipelines on STATE-Bench's enterprise workflows alongside the stateful conversational tool-use tasks in ToolSandbox and ToolTalk in Table~\ref{tab:methodresults}. Four additional static benchmarks assess methods compatible with their retrieval, document-QA, or summarization interfaces, as detailed in Supplement~\ref{app:auxiliary-four}.

\textsc{AgentInputLab} supports local Transformers inference~\citep{wolf2020transformers}, vLLM~\citep{kwon2023vllm} and other OpenAI-compatible servers, as well as native provider APIs. Our experiments span four locally served backbones across a parameter-scale gradient: \code{Qwen3-4B}~\citep{qwen2025qwen3} (Q4), \code{Gemma4-12B} (G12), \code{GPT-OSS-20B} (O20; MXFP4), and \code{Qwen3.8-27B} (Q27; FP8), supplemented in the main table by \code{DeepSeek-V4.1-flash} (version 10-17; DS). This yields $23\times3\times5=345$ method--arena--model configurations and 15 reference entries, executed with prespecified seed 17 and deterministic decoding where supported. Following the STATE-Bench task inventory~\citep{microsoft2026statebench}, we use 450 task groups, assigning 100 to selector fitting, 50 to calibration, and 300 to primary evaluation, while ToolSandbox and ToolTalk contribute 129 and 78 evaluation groups, respectively. Whereas Table~\ref{tab:methodresults} reports a standard \code{OPT0}
reference per arena--backbone setting, the paired analysis in
Section~\ref{sec:rq2} and the fixed-task repeats in
Supplement~\ref{app:component} restore \code{OPT0} to each method's
exact starting state and average within-pair score differences. Model revisions, prompts, tokenizer and chat-template hashes,
reasoning settings, dependencies, hardware, and resource budgets
are recorded in experiment manifests. Across all three arenas, main-table and rank analyses cover all five backbones, paired and routing analyses evaluate five family representatives on Q4 and O20, and the intensity study evaluates a separate five-method set with adjustable parameters listed in Table~\ref{tab:intensity-settings} of Supplement~\ref{app:fair}.

\noindent\textbf{Fair comparison and inference.}
Our local inference resources included eight NVIDIA L20 GPUs, while the hosted backbone used its provider API. Intervention profiles use a standard budget of up to 4,096 tokens and eight calls. 
In paired comparisons, the intervention and \code{OPT0} runs share the
same starting state, backbone, seed, tools, and scorer. Each pipeline
runs under its declared resource profile.
Supplement~\ref{app:fair} provides the full execution and accounting details.

Official captures the task outcome, allowing a successful fallback to earn credit even when an intervention fails. Valid, written as $100Y^T$, preserves that credit only when the intervention finishes and recorded checks confirm no use of privileged information, following Equation~\ref{eq:estimand}.
A graded downstream truncation retains its native score under the same rule. Aggregate scores map to a 0--100 scale while preserving each arena's native criterion, including ToolSandbox's continuous score, with all method comparisons performed within each arena. Confirmed failed runs without an accepted native score are assigned zero credit when computing aggregates over all planned runs, rather than being excluded from the denominator. Unless noted otherwise, interactive-arena gains, rankings, and routing targets rely on Valid.
In the paired analysis, we evaluate one eligible pipeline per family across five operations on restorable task states: InfoCQR rewrites requests, Plan-and-Solve organizes steps, Q-RAG retrieves evidence, Ask or Assume? seeks clarification, and SkillOpt provides skill context. Running \code{OPT0} from each pipeline's initial state produces 2,535 paired comparisons over 507 task groups per backbone on Q4 and O20. Fig.~\ref{fig:alignment-supplement} partitions these eligible inputs by pre-action need into Matched cases, where the input shows the targeted need such as missing user information for Ask or Assume?, and Off-target cases for remaining eligible inputs. The same criteria hold across backbones, and subgroup gains average equally across all three arenas. Pointwise 95\% intervals in Fig.~\ref{fig:motivation}(b)
and Fig.~\ref{fig:alignment-supplement} use 2,000 task-group
bootstrap draws. Routing and feature-ablation intervals in
Fig.~\ref{fig:selector}, and benchmark-heldout intervals in
Fig.~\ref{fig:retrospective-loso}, use 5,000 draws conditional
on the fitted policies. Aggregate scores are reported for the intensity analysis, while Supplement~\ref{app:rq-protocols} details the grouping and resampling procedures.

\noindent\textbf{Selector policy and validation.}
Selector analyses rely on Valid scores, the five family representatives, and disjoint task-group splits, while the benchmark-heldout study refits the same policies and feature representation as detailed in Section~\ref{app:selector-extension}. 
Routing uses task-group outcome tables constructed from the
paired-analysis runs, separately from the main-table cohort,
with \code{OPT0} aggregation defined in
Supplement~\ref{app:selector}. In ToolSandbox, selection is
evaluated offline over recorded pipeline outcomes from
method-specific restored environments grouped by task identity.
Fig.~\ref{fig:selector} compares Threshold rule, Direct ridge, Ridge T, Logistic T, GBDT T, KNN, and Guardrail. These models pair standard linear, logistic, nearest-neighbor, and tree-based learners~\citep{hastie2009,friedman2001} with per-action outcome modeling under the T-learner framework~\citep{kunzel2019}, alongside Threshold rule and Guardrail as rule-based and cost-aware baselines. Before execution, each policy chooses among the five representative pipelines and \code{OPT0} from 14 observable channels organized into seven blocks: request length, request form, constraint anchors, entity and numeric anchors, ambiguity signals, planning signals, and evidence demand, documented in Table~\ref{tab:figure6-features} of Supplement~\ref{app:selector}. These pre-action counts and lexical proxies strictly omit model and arena identities, gold answers, future states, and outcome signals.

In Fig.~\ref{fig:selector}, policies are trained on 100 STATE-Bench development groups, calibrated on 50 held-out groups, and evaluated across all three arenas, with ToolSandbox and ToolTalk used strictly for evaluation. Separate splits for the benchmark-heldout study appear in Table~\ref{tab:heldout-splits}. Three reference policies contextualize routing performance: Random-five selects \code{OPT0} at the fitted policy's rate and distributes the remaining probability uniformly across the five methods, a fixed-action reference applies a single development-selected action to every test task, and the Oracle picks the highest-scoring recorded action on each task. Feature ablations drop an entire feature block before refitting and recalibrating the same learner, and Supplement~\ref{app:selector} specifies normalization, controls, and the paired bootstrap procedure.
\subsection{Main Results}
\label{sec:results}
We report results for the five research questions in sequence.

\begin{table}[!t]
\centering
\caption{Each cell shows smaller gray Official scores above larger black Valid scores. Amber intensity indicates the gap. Model abbreviations follow Section~\ref{sec:exp:setup}.}
\label{tab:methodresults}
\setlength{\tabcolsep}{1.15pt}
\renewcommand{\arraystretch}{0.91}
\resizebox{\columnwidth}{!}{
\begin{tabular}{llrrrrrrrrrrrrrrr}
\toprule
 & & \multicolumn{5}{c}{STATE-Bench} & \multicolumn{5}{c}{ToolSandbox} & \multicolumn{5}{c}{ToolTalk} \\
\cmidrule(lr){3-7}\cmidrule(lr){8-12}\cmidrule(lr){13-17}
Family & Method & Q4 & G12 & O20 & Q27 & DS & Q4 & G12 & O20 & Q27 & DS & Q4 & G12 & O20 & Q27 & DS \\
\midrule
B & \code{OPT0}~\cite{yao2023react} & \scorepair{4.0}{4.0}{0} & \scorepair{11.0}{11.0}{0} & \scorepair{7.3}{7.3}{0} & \scorepair{12.3}{12.3}{0} & \scorepair{13.0}{13.0}{0} & \scorepair{22.4}{22.4}{0} & \scorepair{25.9}{25.9}{0} & \scorepair{19.7}{19.7}{0} & \scorepair{22.1}{22.1}{0} & \scorepair{63.1}{63.1}{0} & \scorepair{28.2}{28.2}{0} & \scorepair{37.2}{37.2}{0} & \scorepair{38.5}{38.5}{0} & \scorepair{39.7}{39.7}{0}  & \scorepair{42.3}{42.3}{0} \\
\midrule
R & InfoCQR~\cite{ye2023infocqr} & \scorepair{12.0}{12.0}{0} & \scorepair{29.0}{29.0}{0} & \scorepair{22.0}{21.3}{1} & \scorepair{30.7}{30.7}{0} & \scorepair{38.7}{38.7}{0} & \scorepair{21.5}{21.5}{0} & \scorepair{26.2}{26.2}{0} & \scorepair{14.9}{14.5}{1} & \scorepair{25.9}{25.9}{0} & \scorepair{70.7}{70.7}{0} & \scorepair{28.2}{28.2}{0} & \scorepair{52.6}{52.6}{0} & \scorepair{50.0}{48.7}{2} & \scorepair{46.2}{46.2}{0}  & \scorepair{56.4}{56.4}{0} \\
 & Query2doc~\cite{wang2023query2doc} & \scorepair{14.3}{14.3}{0} & \scorepair{26.0}{26.0}{0} & \scorepair{22.7}{22.7}{0} & \scorepair{29.0}{29.0}{0} & \scorepair{39.3}{39.3}{0} & \scorepair{20.7}{20.7}{0} & \scorepair{27.6}{27.6}{0} & \scorepair{14.2}{14.2}{0} & \scorepair{26.1}{26.1}{0} & \scorepair{74.5}{74.5}{0} & \scorepair{23.1}{23.1}{0} & \scorepair{52.6}{52.6}{0} & \scorepair{50.0}{50.0}{0} & \scorepair{56.4}{56.4}{0}  & \scorepair{60.3}{60.3}{0} \\
 & RRR~\cite{ma2023queryrewriting} & \scorepair{13.0}{12.7}{0} & \scorepair{25.3}{25.3}{0} & \scorepair{20.7}{19.7}{1} & \scorepair{30.3}{30.3}{0} & \scorepair{37.3}{36.0}{2} & \scorepair{19.2}{19.2}{0} & \scorepair{24.8}{24.8}{0} & \scorepair{22.1}{18.0}{5} & \scorepair{23.6}{23.6}{0} & \scorepair{66.1}{66.1}{0} & \scorepair{24.4}{24.4}{0} & \scorepair{37.2}{37.2}{0} & \scorepair{35.9}{26.9}{12} & \scorepair{37.2}{37.2}{0}  & \scorepair{34.6}{34.6}{0} \\
 & ReFormeR~\cite{bigdeli2026reformer} & \scorepair{10.3}{10.3}{0} & \scorepair{20.3}{20.3}{0} & \scorepair{15.3}{8.0}{10} & \scorepair{25.0}{25.0}{0} & \scorepair{30.0}{30.0}{0} & \scorepair{20.7}{18.2}{3} & \scorepair{27.4}{27.4}{0} & \scorepair{23.4}{14.4}{12} & \scorepair{25.1}{25.1}{0} & \scorepair{72.7}{72.7}{0} & \scorepair{26.9}{26.9}{0} & \scorepair{50.0}{50.0}{0} & \scorepair{38.5}{23.1}{21} & \scorepair{48.7}{48.7}{0}  & \scorepair{55.1}{55.1}{0} \\
\midrule
D & Plan-and-Solve~\cite{wang2023planandsolve} & \scorepair{13.0}{13.0}{0} & \scorepair{32.7}{32.7}{0} & \scorepair{23.3}{23.3}{0} & \scorepair{30.3}{30.3}{0} & \scorepair{43.0}{43.0}{0} & \scorepair{20.7}{20.7}{0} & \scorepair{24.1}{24.1}{0} & \scorepair{12.0}{12.0}{0} & \scorepair{23.2}{23.2}{0} & \scorepair{74.6}{74.6}{0} & \scorepair{33.3}{33.3}{0} & \scorepair{57.7}{57.7}{0} & \scorepair{44.9}{44.9}{0} & \scorepair{60.3}{60.3}{0}  & \scorepair{66.7}{66.7}{0} \\
 & Decomp~\cite{khot2023decomposed} & \scorepair{10.7}{5.3}{7} & \scorepair{14.7}{13.3}{2} & \scorepair{18.7}{10.7}{11} & \scorepair{12.3}{10.3}{3} & \scorepair{27.0}{27.0}{0} & \scorepair{20.7}{14.2}{9} & \scorepair{24.7}{7.4}{23} & \scorepair{24.9}{12.4}{17} & \scorepair{24.3}{15.7}{11} & \scorepair{69.2}{67.3}{3} & \scorepair{30.8}{15.4}{21} & \scorepair{43.6}{17.9}{34} & \scorepair{42.3}{6.4}{48} & \scorepair{37.2}{20.5}{22}  & \scorepair{47.4}{47.4}{0} \\
 & Self-Ask~\cite{press2023selfask} & \scorepair{10.3}{10.0}{0} & \scorepair{14.7}{14.7}{0} & \scorepair{16.3}{4.3}{16} & \scorepair{25.0}{25.0}{0} & \scorepair{14.7}{14.7}{0} & \scorepair{19.2}{19.2}{0} & \scorepair{25.1}{16.2}{12} & \scorepair{13.1}{8.5}{6} & \scorepair{23.9}{23.9}{0} & \scorepair{64.8}{64.8}{0} & \scorepair{20.5}{20.5}{0} & \scorepair{34.6}{34.6}{0} & \scorepair{34.6}{19.2}{21} & \scorepair{52.6}{52.6}{0}  & \scorepair{38.5}{38.5}{0} \\
 & LLMCompiler~\cite{kim2024llmcompiler} & \scorepair{13.0}{13.0}{0} & \scorepair{28.3}{28.3}{0} & \scorepair{17.3}{1.3}{21} & \scorepair{29.3}{28.7}{1} & \scorepair{38.7}{38.7}{0} & \scorepair{16.1}{15.6}{1} & \scorepair{25.5}{20.8}{6} & \scorepair{23.7}{3.1}{27} & \scorepair{22.3}{20.7}{2} & \scorepair{72.4}{69.3}{4} & \scorepair{28.2}{21.8}{9} & \scorepair{52.6}{52.6}{0} & \scorepair{48.7}{21.8}{36} & \scorepair{47.4}{42.3}{7}  & \scorepair{53.8}{53.8}{0} \\
 & Plan-and-Budget~\cite{lin2026planbudget} & \scorepair{12.3}{12.3}{0} & \scorepair{19.7}{19.7}{0} & \scorepair{12.3}{11.0}{2} & \scorepair{25.0}{25.0}{0} & \scorepair{33.0}{33.0}{0} & \scorepair{19.2}{17.6}{2} & \scorepair{24.6}{24.6}{0} & \scorepair{13.6}{13.4}{0} & \scorepair{24.7}{24.7}{0} & \scorepair{73.6}{73.6}{0} & \scorepair{30.8}{30.8}{0} & \scorepair{51.3}{51.3}{0} & \scorepair{38.5}{37.2}{2} & \scorepair{50.0}{50.0}{0}  & \scorepair{50.0}{50.0}{0} \\
\midrule
A & Q-RAG~\cite{sorokin2026qrag} & \scorepair{13.7}{13.7}{0} & \scorepair{35.0}{35.0}{0} & \scorepair{26.0}{26.0}{0} & \scorepair{35.3}{35.3}{0} & \scorepair{49.7}{49.7}{0} & \scorepair{17.6}{17.6}{0} & \scorepair{23.1}{23.1}{0} & \scorepair{8.9}{8.9}{0} & \scorepair{22.1}{22.1}{0} & \scorepair{69.9}{69.9}{0} & \scorepair{30.8}{30.8}{0} & \scorepair{59.0}{59.0}{0} & \scorepair{51.3}{51.3}{0} & \scorepair{48.7}{48.7}{0}  & \scorepair{65.4}{65.4}{0} \\
 & IRCoT~\cite{trivedi2023ircot} & \scorepair{14.3}{14.3}{0} & \scorepair{14.7}{14.7}{0} & \scorepair{18.3}{11.0}{10} & \scorepair{15.7}{15.7}{0} & \scorepair{22.0}{22.0}{0} & \scorepair{19.2}{19.2}{0} & \scorepair{22.8}{12.2}{14} & \scorepair{23.5}{6.2}{23} & \scorepair{23.5}{23.5}{0} & \scorepair{58.7}{58.7}{0} & \scorepair{24.4}{24.4}{0} & \scorepair{38.5}{38.5}{0} & \scorepair{38.5}{24.4}{19} & \scorepair{38.5}{38.5}{0}  & \scorepair{44.9}{44.9}{0} \\
 & Adaptive-RAG~\cite{jeong2024adaptiverag} & \scorepair{15.0}{15.0}{0} & \scorepair{30.0}{30.0}{0} & \scorepair{27.0}{25.7}{2} & \scorepair{33.0}{33.0}{0} & \scorepair{44.3}{44.3}{0} & \scorepair{18.4}{18.4}{0} & \scorepair{22.8}{21.2}{2} & \scorepair{13.4}{10.9}{3} & \scorepair{21.8}{21.8}{0} & \scorepair{66.9}{66.9}{0} & \scorepair{26.9}{26.9}{0} & \scorepair{56.4}{56.4}{0} & \scorepair{48.7}{48.7}{0} & \scorepair{53.8}{53.8}{0}  & \scorepair{62.8}{62.8}{0} \\
 & $R^2$AG~\cite{ye2024r2ag} & \scorepair{13.7}{10.7}{4} & \scorepair{27.3}{21.3}{8} & \scorepair{22.0}{17.3}{6} & \scorepair{31.3}{26.0}{7} & \scorepair{47.0}{41.7}{7} & \scorepair{18.4}{18.4}{0} & \scorepair{22.1}{18.8}{4} & \scorepair{11.3}{11.3}{0} & \scorepair{23.3}{23.3}{0} & \scorepair{66.6}{66.6}{0} & \scorepair{35.9}{35.9}{0} & \scorepair{59.0}{57.7}{2} & \scorepair{47.4}{47.4}{0} & \scorepair{55.1}{55.1}{0}  & \scorepair{62.8}{62.8}{0} \\
\midrule
C & Ask or Assume?~\cite{edwards2026askassume} & \scorepair{14.7}{4.0}{14} & \scorepair{29.7}{29.7}{0} & \scorepair{17.3}{13.7}{5} & \scorepair{29.3}{17.7}{16} & \scorepair{40.7}{40.3}{1} & \scorepair{22.6}{17.1}{7} & \scorepair{26.0}{26.0}{0} & \scorepair{13.5}{13.5}{0} & \scorepair{24.5}{11.0}{18} & \scorepair{75.9}{74.5}{2} & \scorepair{29.5}{25.6}{5} & \scorepair{48.7}{46.2}{3} & \scorepair{34.6}{34.6}{0} & \scorepair{57.7}{42.3}{21}  & \scorepair{53.8}{53.8}{0} \\
 & UDCS~\cite{matsnev2026udcs} & \scorepair{13.7}{13.7}{0} & \scorepair{30.0}{30.0}{0} & \scorepair{21.0}{10.7}{14} & \scorepair{30.3}{7.0}{31} & \scorepair{42.7}{42.0}{1} & \scorepair{24.6}{24.6}{0} & \scorepair{27.0}{27.0}{0} & \scorepair{11.6}{9.8}{2} & \scorepair{25.2}{6.2}{25} & \scorepair{74.5}{72.4}{3} & \scorepair{19.2}{19.2}{0} & \scorepair{60.3}{60.3}{0} & \scorepair{41.0}{39.7}{2} & \scorepair{57.7}{30.8}{36}  & \scorepair{56.4}{56.4}{0} \\
 & ATD~\cite{kobalczyk2025active} & \scorepair{11.7}{11.7}{0} & \scorepair{7.7}{7.7}{0} & \scorepair{2.7}{2.7}{0} & \scorepair{31.0}{31.0}{0} & \scorepair{36.7}{36.7}{0} & \scorepair{25.9}{25.9}{0} & \scorepair{7.7}{7.7}{0} & \scorepair{11.2}{11.2}{0} & \scorepair{20.4}{20.4}{0} & \scorepair{69.8}{69.8}{0} & \scorepair{12.8}{12.8}{0} & \scorepair{1.3}{1.3}{0} & \scorepair{6.4}{6.4}{0} & \scorepair{12.8}{12.8}{0}  & \scorepair{17.9}{17.9}{0} \\
\midrule
U & SkillOpt~\cite{yang2026skillopt} & \scorepair{12.7}{12.7}{0} & \scorepair{32.0}{32.0}{0} & \scorepair{23.3}{23.3}{0} & \scorepair{32.7}{32.7}{0} & \scorepair{47.3}{47.3}{0} & \scorepair{20.0}{20.0}{0} & \scorepair{26.0}{26.0}{0} & \scorepair{13.3}{13.3}{0} & \scorepair{25.1}{25.1}{0} & \scorepair{77.6}{77.6}{0} & \scorepair{30.8}{30.8}{0} & \scorepair{59.0}{59.0}{0} & \scorepair{50.0}{50.0}{0} & \scorepair{60.3}{60.3}{0}  & \scorepair{70.5}{70.5}{0} \\
 & LongLLMLingua~\cite{jiang2024longllmlingua} & \scorepair{12.3}{12.3}{0} & \scorepair{32.0}{32.0}{0} & \scorepair{21.3}{21.3}{0} & \scorepair{32.7}{32.7}{0} & \scorepair{45.7}{45.7}{0} & \scorepair{20.7}{20.7}{0} & \scorepair{25.3}{25.3}{0} & \scorepair{10.6}{10.6}{0} & \scorepair{24.5}{24.5}{0} & \scorepair{73.0}{73.0}{0} & \scorepair{26.9}{26.9}{0} & \scorepair{61.5}{61.5}{0} & \scorepair{46.2}{46.2}{0} & \scorepair{50.0}{50.0}{0}  & \scorepair{60.3}{60.3}{0} \\
 & LLMLingua-2~\cite{pan2024llmlingua2} & \scorepair{12.3}{12.3}{0} & \scorepair{35.0}{35.0}{0} & \scorepair{23.7}{23.7}{0} & \scorepair{33.3}{33.3}{0} & \scorepair{40.0}{40.0}{0} & \scorepair{21.5}{21.5}{0} & \scorepair{26.5}{26.5}{0} & \scorepair{12.0}{12.0}{0} & \scorepair{25.1}{25.1}{0} & \scorepair{72.9}{72.9}{0} & \scorepair{30.8}{30.8}{0} & \scorepair{60.3}{60.3}{0} & \scorepair{47.4}{47.4}{0} & \scorepair{56.4}{56.4}{0}  & \scorepair{61.5}{61.5}{0} \\
 & RECOMP~\cite{xu2024recomp} & \scorepair{11.7}{11.7}{0} & \scorepair{31.7}{31.7}{0} & \scorepair{23.0}{23.0}{0} & \scorepair{28.0}{28.0}{0} & \scorepair{41.7}{41.7}{0} & \scorepair{20.7}{20.7}{0} & \scorepair{26.3}{26.3}{0} & \scorepair{23.0}{23.0}{0} & \scorepair{24.7}{24.7}{0} & \scorepair{74.0}{74.0}{0} & \scorepair{29.5}{29.5}{0} & \scorepair{55.1}{55.1}{0} & \scorepair{46.2}{46.2}{0} & \scorepair{52.6}{52.6}{0}  & \scorepair{61.5}{61.5}{0} \\
 & ACON~\cite{kang2026acon} & \scorepair{14.3}{14.3}{0} & \scorepair{27.7}{27.7}{0} & \scorepair{20.7}{20.7}{0} & \scorepair{28.3}{28.3}{0} & \scorepair{41.0}{41.0}{0} & \scorepair{20.7}{20.7}{0} & \scorepair{26.9}{26.9}{0} & \scorepair{14.1}{13.5}{1} & \scorepair{24.6}{24.6}{0} & \scorepair{77.7}{77.7}{0} & \scorepair{28.2}{28.2}{0} & \scorepair{61.5}{61.5}{0} & \scorepair{50.0}{35.9}{19} & \scorepair{57.7}{57.7}{0}  & \scorepair{57.7}{57.7}{0} \\
 & ACE~\cite{zhang2026ace} & \scorepair{13.0}{13.0}{0} & \scorepair{32.7}{32.7}{0} & \scorepair{25.0}{25.0}{0} & \scorepair{32.7}{32.7}{0} & \scorepair{43.7}{43.7}{0} & \scorepair{19.2}{19.2}{0} & \scorepair{26.2}{26.2}{0} & \scorepair{13.4}{13.4}{0} & \scorepair{23.6}{23.6}{0} & \scorepair{74.3}{74.3}{0} & \scorepair{32.1}{32.1}{0} & \scorepair{60.3}{60.3}{0} & \scorepair{50.0}{50.0}{0} & \scorepair{61.5}{61.5}{0}  & \scorepair{70.5}{70.5}{0} \\
 & MCE~\cite{ye2026mce} & \scorepair{13.0}{13.0}{0} & \scorepair{33.3}{33.3}{0} & \scorepair{22.0}{22.0}{0} & \scorepair{33.3}{33.3}{0} & \scorepair{47.0}{47.0}{0} & \scorepair{21.5}{21.5}{0} & \scorepair{26.3}{26.3}{0} & \scorepair{14.0}{14.0}{0} & \scorepair{25.0}{25.0}{0} & \scorepair{74.6}{74.6}{0} & \scorepair{30.8}{30.8}{0} & \scorepair{56.4}{56.4}{0} & \scorepair{47.4}{47.4}{0} & \scorepair{60.3}{60.3}{0}  & \scorepair{64.1}{64.1}{0} \\
\bottomrule
\end{tabular}
}
\end{table}

\subsubsection{RQ1: Method rankings depend on the arena and backbone}
\label{sec:rq1}
This evaluation tests whether a method chosen in one setting stays effective when the task environment or backbone changes. In Table~\ref{tab:methodresults}, acquisition shows a recurring advantage on STATE-Bench, leading or tying across every backbone, while conversational tool-use arenas exhibit a wider set of leading families. DeepSeek-V4.1-flash highlights this shift: Q-RAG leads on STATE-Bench, ACON leads on ToolSandbox, and ACE and SkillOpt tie on ToolTalk, with its \code{OPT0} also far surpassing local-model baselines on ToolSandbox. While the hosted DS backbone reaches higher peak scores in every arena, changing leaders confirm that pipeline choice depends heavily on the task.

Reading these rankings also calls for distinguishing task outcome from intervention completion. For ToolTalk under O20, Decomp reaches 42.3 Official but only 6.4 Valid because a downstream agent can still fulfill the task after decomposition reports failure, leaving task credit in Official that Valid strips away. Deleting a meeting successfully, for instance, can still earn task credit despite an earlier decomposition failure, whereas Q-RAG achieves 51.3 on both metrics. 
Table~\ref{tab:score-evidence} compares Official and Valid on the
same natively scored tasks for each method, with equal weight across
methods. The largest average gap occurs on O20--ToolTalk.
On O20--ToolSandbox, 21 methods score above the standard \code{OPT0}
reference on their shared scored-task subsets under Official, compared
with 13 under Valid.
These comparisons isolate the effect of the scoring rule on observed
outcomes; Table~\ref{tab:methodresults} reports performance over all
planned tasks. Supplement~\ref{app:retrieval-archive} examines the
corresponding full-task rankings.

\begin{table}[t]
\centering\scriptsize
\setlength{\tabcolsep}{3pt}
\caption{Official and Valid scores on the same natively scored tasks
for each method, averaged equally over the 23 methods.
The final column counts methods outperforming OPT0 on their shared
scored-task subsets. S, X, and T denote STATE-Bench, ToolSandbox,
and ToolTalk.}
\label{tab:score-evidence}
\begin{tabular}{@{}llrrr@{}}
\toprule
& & \multicolumn{2}{c}{Average score (0--100)} & Above \code{OPT0} \\
\cmidrule(lr){3-4}
Model & Arena & Official & Valid & O / V (of 23) \\
\midrule
Q4 & S & 12.88 & 12.03 & 23 / 22 \\
Q4 & X & 20.63 & 19.91 & 3 / 2 \\
Q4 & T & 27.61 & 26.49 & 11 / 9 \\
\addlinespace[2pt]
G12 & S & 27.32 & 27.00 & 23 / 23 \\
G12 & X & 25.48 & 23.46 & 11 / 11 \\
G12 & T & 51.14 & 49.86 & 20 / 19 \\
\addlinespace[2pt]
O20 & S & 20.95 & 17.66 & 23 / 21 \\
O20 & X & 22.26 & 18.64 & 21 / 13 \\
O20 & T & 43.66 & 37.76 & 16 / 13 \\
\addlinespace[2pt]
Q27 & S & 28.95 & 27.08 & 22 / 21 \\
Q27 & X & 24.20 & 22.34 & 22 / 18 \\
Q27 & T & 50.58 & 47.79 & 19 / 18 \\
\addlinespace[2pt]
DS & S & 38.77 & 38.44 & 23 / 23 \\
DS & X & 71.65 & 71.28 & 22 / 22 \\
DS & T & 55.35 & 55.35 & 20 / 20 \\
\addlinespace[2pt]
\bottomrule
\end{tabular}
\end{table}

Task dependence extends beyond the interactive arenas as well. Under DeepSeek-V4.1-flash in Table~\ref{tab:auxiliary-four} of Supplement~\ref{app:auxiliary-four}, Query2doc leads on BRIGHT Earth Science, FiQA-dev, and MultiFieldQA-en, whereas RRR achieves the highest observed ROUGE-L on QMSum. Plan-and-Budget's point scores sit above \code{OPT0} on BRIGHT and document-QA benchmarks but below it on FiQA and QMSum.  The choice of input preparation therefore depends on the task,
output requirements, and system configuration.

\begin{figure}[t]
    \centering

    {\footnotesize\mbox{\textcolor[HTML]{255278}{\rule{1em}{0.65em}}
        ~Higher
        \hspace{.6em}
        \textcolor[HTML]{E6EBEF}{\rule{1em}{0.65em}}
        ~Equal
        \hspace{.6em}
        \textcolor[HTML]{AD7830}{\rule{1em}{0.65em}}
        ~Lower}}

    \par\medskip

    \subfigure[STATE-Bench.\label{fig:effects-statebench}]{%
        \includegraphics[width=0.305\columnwidth]{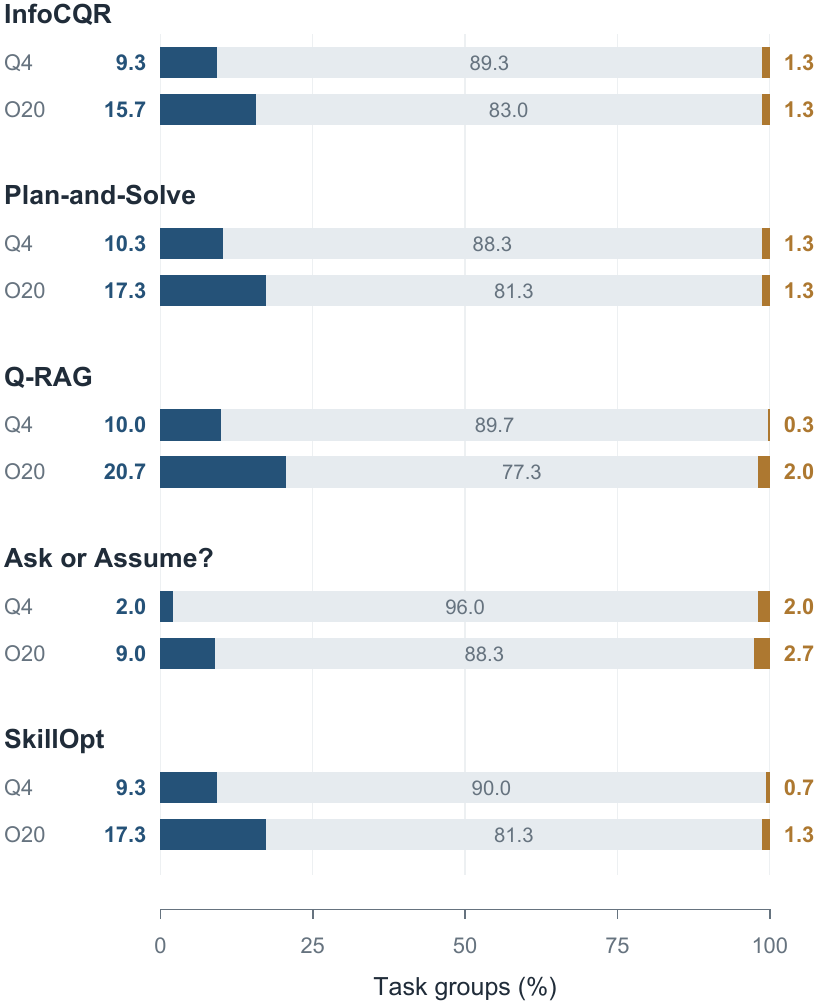}}%
    \hfill%
    \subfigure[ToolSandbox.\label{fig:effects-toolsandbox}]{%
        \includegraphics[width=0.305\columnwidth]{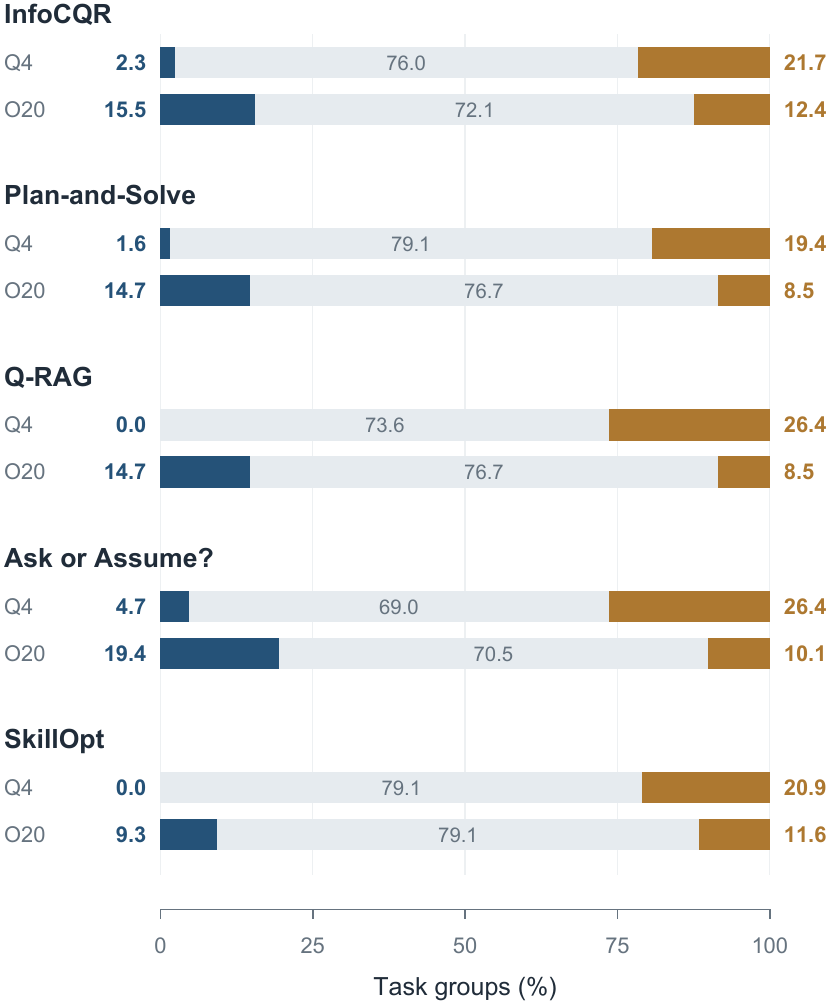}}%
    \hfill%
    \subfigure[ToolTalk.\label{fig:effects-tooltalk}]{%
        \includegraphics[width=0.305\columnwidth]{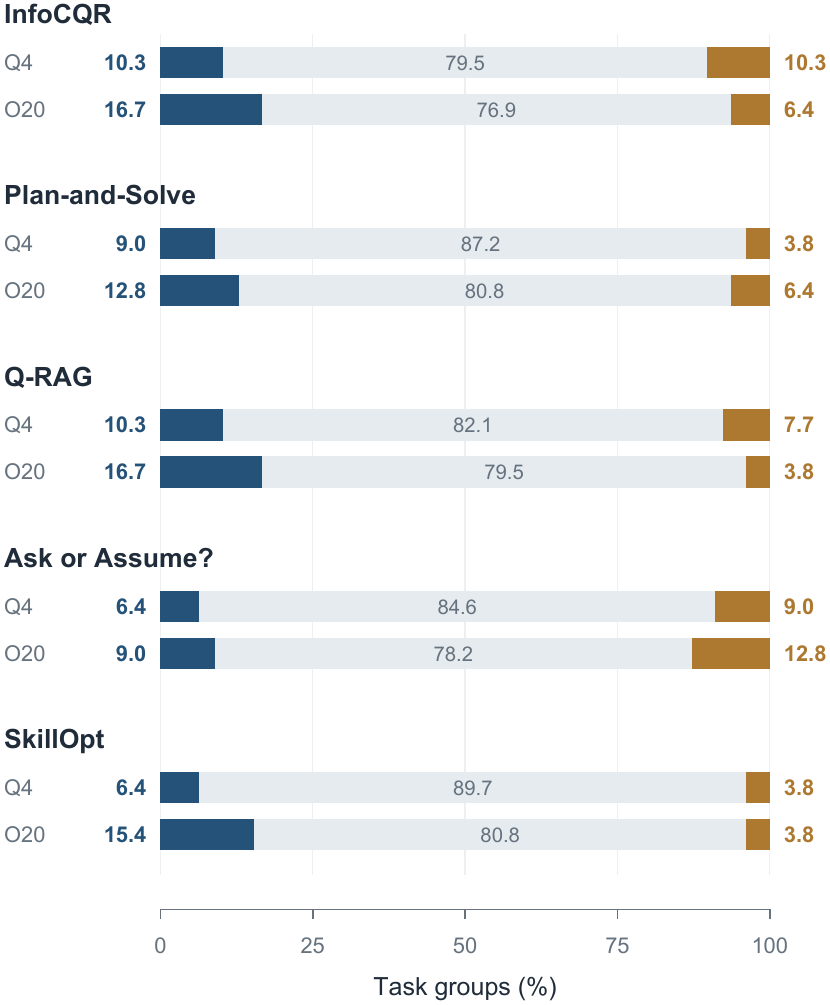}}

    \caption{\textbf{Task-level gains and losses relative to \code{OPT0}.} Blue, gray, and gold denote higher, equal, and lower Valid scores
relative to the paired \code{OPT0} run from the same initial state. Bars retain failed tasks; labels are percentages computed before rounding. Q4 and O20 are shown separately.}
    \label{fig:effects}

    \Description{
        Three panels show stacked horizontal bars for five
        representative pipelines and two backbones.
        Each bar separates the proportions of task groups with
        higher, equal, or lower valid scores than \code{OPT0}.
    }
\end{figure}

\subsubsection{RQ2: Benefits depend on the observed input and support set}
\label{sec:rq2}
Task-level differences show whether an average gain is broadly shared or combines benefits on some inputs with harm on others. Fig.~\ref{fig:effects} shows broad improvements for the five representatives on STATE-Bench alongside a very different balance on ToolSandbox. On Q4, paired mean differences keep all five pipelines below their \code{OPT0} references on ToolSandbox while remaining nonnegative on STATE-Bench. This reversal matters more than a minor rank shift: a pipeline that helps a workflow can still disrupt a stateful conversational task. Fixed-task repeats over a prespecified 30-task subset per setting in Table~\ref{tab:repeat-evidence} of Supplement~\ref{app:component} separate recurring task dependence from execution-sensitive differences. On Q4, InfoCQR and Q-RAG raise STATE-Bench scores and lower ToolTalk point scores across all three runs under both metrics, whereas their gains on ToolSandbox change sign. Across all three runs on O20, all five methods also improve STATE-Bench scores. These consistent directions support task-specific configuration choices, while sign-changing results point to settings where repeated runs are needed to identify a reliable operating point.

\begin{figure}[t]
\centering
\includegraphics[width=\columnwidth]{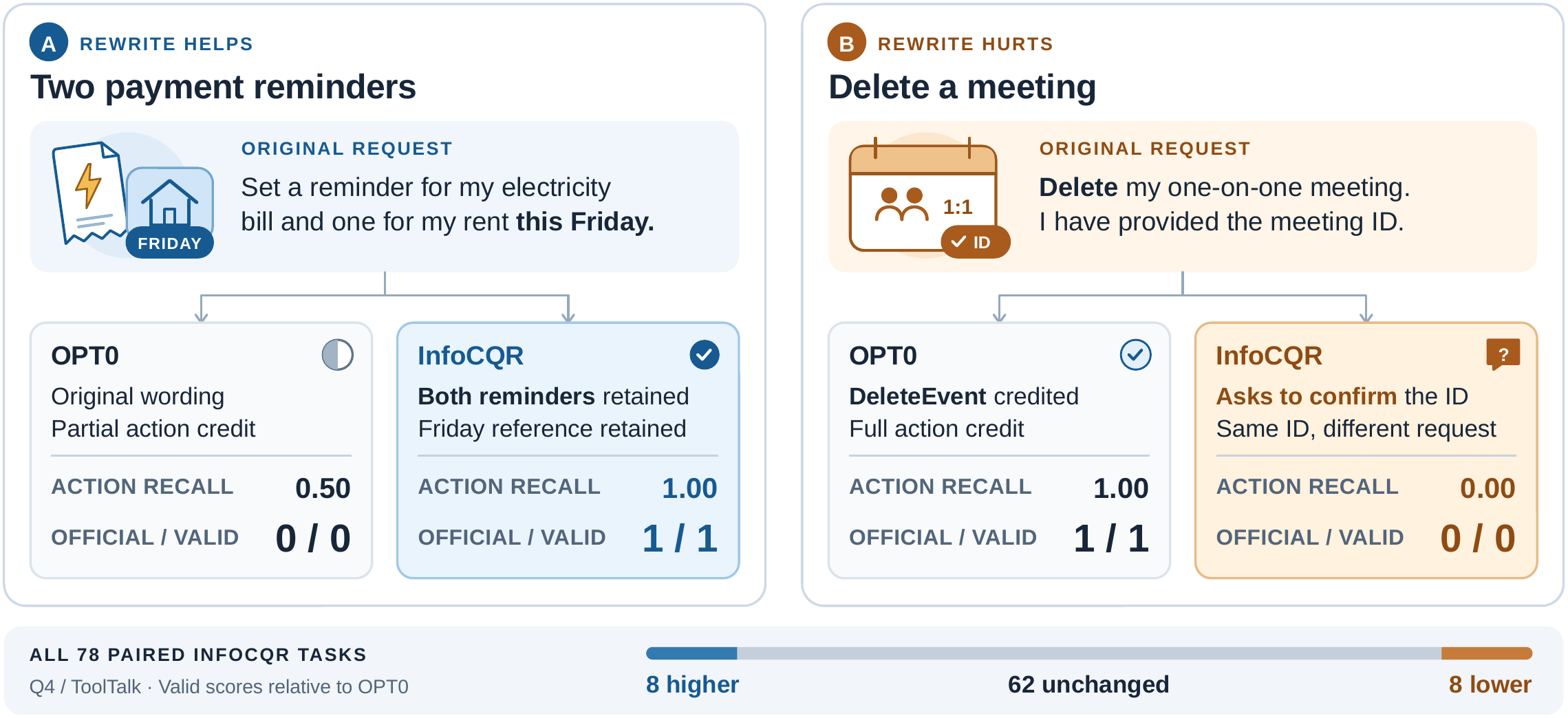}
\caption{\textbf{One rewriter, opposite outcomes.} Two InfoCQR cases on ToolTalk with Q4, each paired with \code{OPT0} from the same restored checkpoint under the declared pipeline profiles. Requests are paraphrased, Official and Valid scores lie on a 0 to 1 scale, and both score differences recur across all three executions of these tasks.}
\Description{Two cases: a rewrite preserves two payment
reminders, while another retains a meeting ID but changes deletion into
confirmation. Action recall increases from 0.50 to 1.00 in the reminder
case and falls from 1.00 to 0.00 in the meeting case. Across 78 paired
InfoCQR tasks, 8 Valid scores increase, 8 decrease, and 62 remain unchanged.}
\label{fig:infocqr-cases}
\end{figure}

A zero average gain can mask opposing effects across individual tasks. Across the full Q4 ToolTalk set, InfoCQR raises 8 of 78 Valid scores, lowers 8, and leaves 62 unchanged, as illustrated in Fig.~\ref{fig:infocqr-cases}. In the reminder case, preserving both payment reminders and the Friday reference lifts native action recall from 0.50 to 1.00, where action recall tracks required-action coverage and partial completion receives no native task-success credit. In the meeting case, keeping the supplied ID while turning a deletion request into a confirmation question passes the lexical contract check yet drops action recall from 1.00 to 0.00. This failure shows that rewrite checks must verify both the requested action and its referenced entities. Across the 5,070 paired method-task comparisons for the five representatives over Q4 and O20 and all three arenas, Valid scores increase in 557 cases, or 11.0\%, decrease in 310 cases, or 6.1\%, and remain unchanged in 4,203 cases, or 82.9\%. These Valid-score losses include both lower task scores after completed interventions and failures to complete the intervention. Paired task scores and execution records help distinguish the two.

\begin{figure}[t]
\centering
\subfigure[Q4.\label{fig:alignment-q4}]{%
\includegraphics[width=.48\columnwidth]{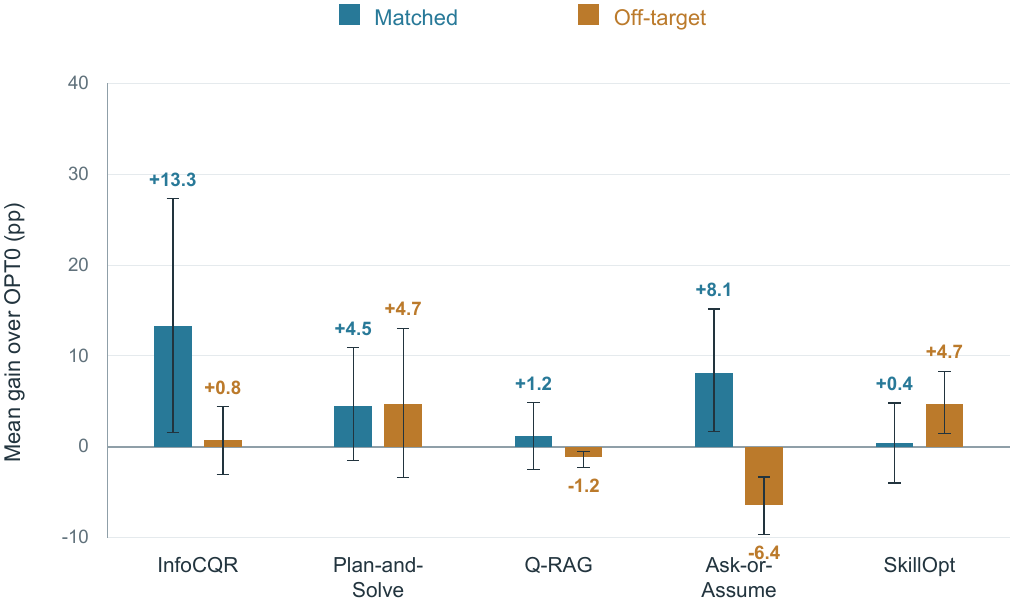}}%
\hfill%
\subfigure[O20.\label{fig:alignment-o20}]{%
\includegraphics[width=.48\columnwidth]{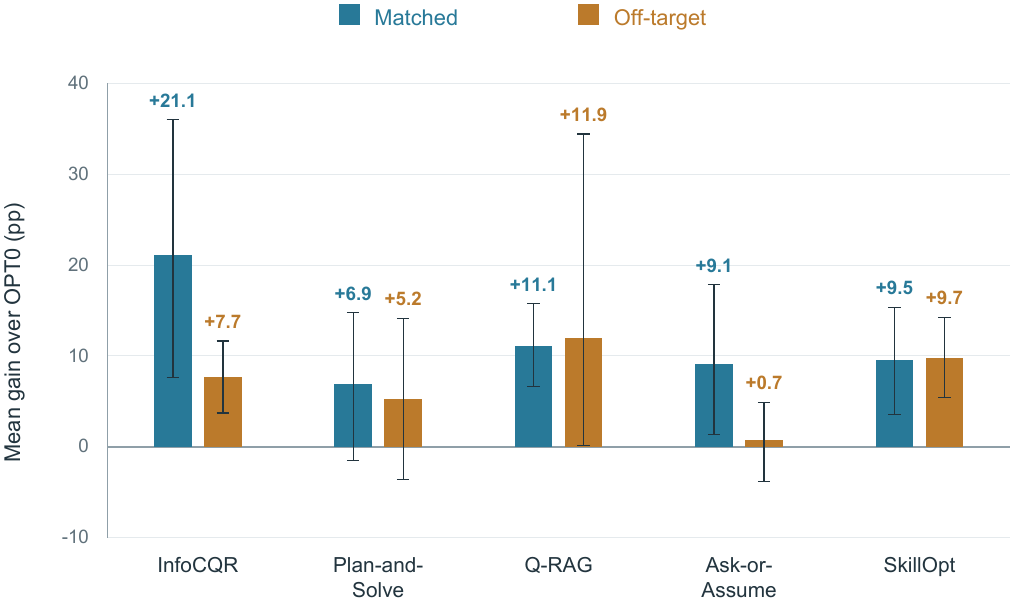}}%
\caption{\textbf{Input-demand gain stratification relative to \code{OPT0}.}
Bars show the equal-weight mean of arena-specific Matched and Off-target
$Y^T$ gains (pp) across STATE-Bench, ToolSandbox, and ToolTalk. Error bars indicate
95\% task-cluster bootstrap intervals.}
\Description{Two grouped-bar panels show Q4 and O20 mean gains over
\code{OPT0} for five methods, stratified by Matched and Off-target demand
and averaged equally across three arenas.}
\label{fig:alignment-supplement}
\end{figure}

While Fig.~\ref{fig:effects} tracks how frequently scores change,
Fig.~\ref{fig:alignment-supplement} compares average gain magnitudes
across input-demand groups using equal arena weights.
Magnitude is essential alongside frequency because ToolSandbox
uses a continuous score. Both InfoCQR and Ask or Assume? attain
higher Matched-group mean gains across both backbones, and
Ask or Assume? shows Matched-minus-Off-target gaps of 14.5 pp
on Q4 and 8.4 pp on O20. Under Q4, Plan-and-Solve is nearly
even across groups with 4.5 pp versus 4.7 pp, whereas SkillOpt
achieves larger gains on Off-target inputs with 4.7 pp versus
0.4 pp. Demand alignment is informative for some interventions
but does not consistently order gains across methods and
backbones. These patterns favor method-specific use of demand
signals rather than treating a nominal demand as sufficient
reason to intervene. Execution traces also help explain why
an apparent demand may not translate into a task gain:
in ToolTalk, asking a clarification question can terminate the
evaluated turn before a tool call. Supplement~\ref{app:component}
provides the component checks.

\begin{figure}[t]
    \centering

    \subfigure[Qwen3-4B (Q4).]{
        \includegraphics[width=.97\columnwidth]{
            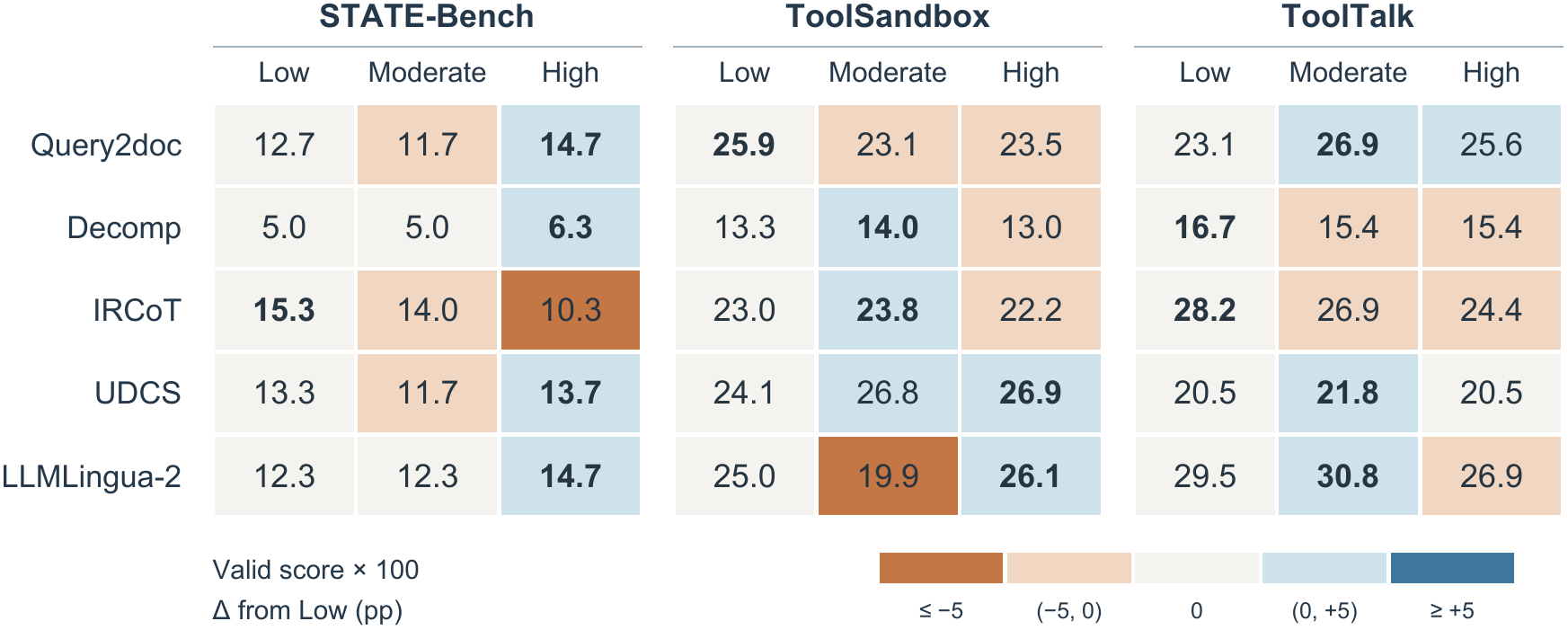
        }
        \label{fig:mechanisms-q4}
    }
    \par\vspace{0.2em}
    \subfigure[GPT-OSS-20B (O20).]{
        \includegraphics[width=.97\columnwidth]{
            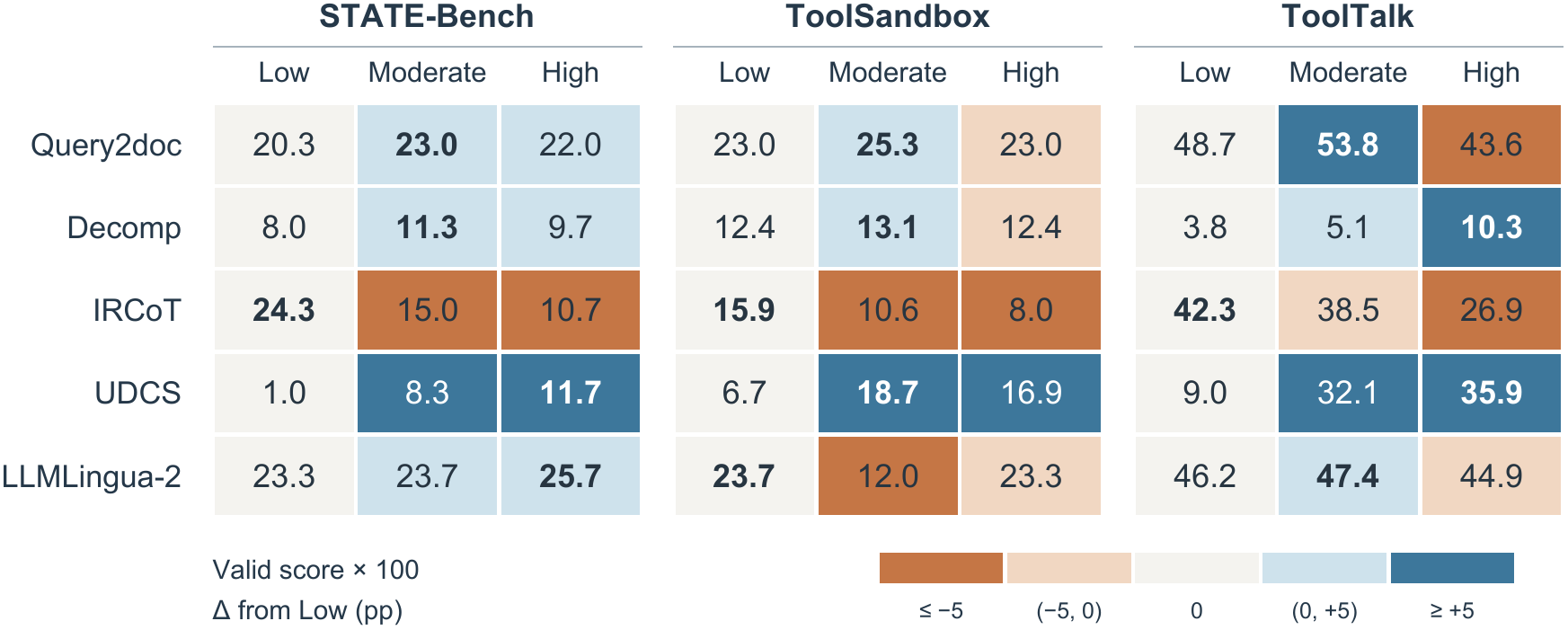
        }
        \label{fig:mechanisms-o20}
    }

    \caption{\textbf{Method-intensity responses for Q4 and O20.} Cells show Valid scores; colors show change from Low. Bold marks the best level, including ties. Levels are method-specific configurations under a fixed total budget.}

    \Description{
        Two heatmaps show five methods across STATE-Bench,
        ToolSandbox, and ToolTalk for Q4 and O20.
        Each arena contains Low, Moderate, and High intensity
        columns. Numbers show valid scores; colors show
        changes from Low, and bold numbers mark the highest
        score within each three-setting group.
    }

    \label{fig:mechanisms}
\end{figure}

\subsubsection{RQ3: Intensity effects are method- and arena-dependent}
This experiment tests whether applying a stronger intervention reliably raises its utility, which shapes how a method is configured after selection. In Fig.~\ref{fig:mechanisms}, responses can move in opposite directions even within the same arena under fixed total budgets. On STATE-Bench, IRCoT scores drop from 15.3 to 14.0 to 10.3 under Q4 and from 24.3 to 15.0 to 10.7 under O20, whereas O20 UDCS rises from 1.0 to 8.3 to 11.7.
Useful operating points also shift across arenas. O20 Decomp peaks at Moderate on STATE-Bench and ToolSandbox with 11.3 and 13.1 points, while peaking at High on ToolTalk with 10.3 points. Query2doc likewise favors an intermediate setting: its O20 STATE-Bench score peaks at Moderate, and the same Moderate peak appears on ToolTalk for both backbones. By contrast, O20 UDCS improves across all three intensity levels on ToolTalk. These patterns support method- and arena-specific calibration rather than a single aggressive default.

\begin{figure}[t]
\centering
\includegraphics[width=\columnwidth]{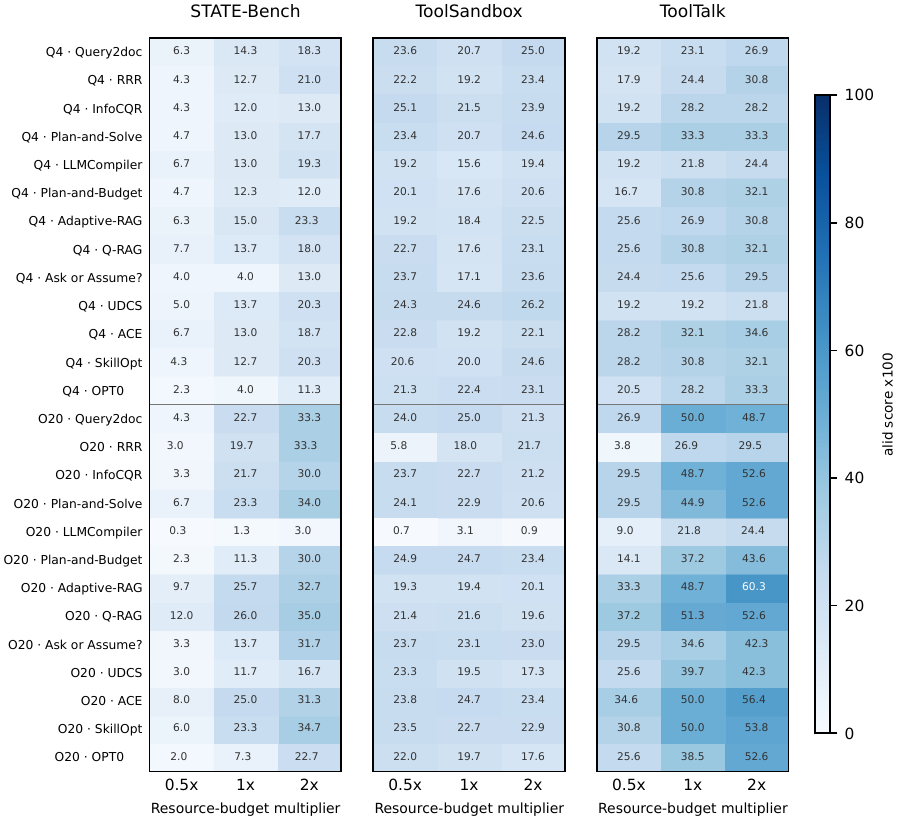}
\caption{\textbf{Resource-budget responses for Q4 and O20.}
Cells show average Valid scores ($100Y^T$) over all planned groups,
including failures, under the respective $0.5\times$, $1\times$, and
$2\times$ profiles. Method-intensity settings remain fixed.}
\Description{Three heatmaps show all 234 average valid scores, with the same
method and model rows in STATE-Bench, ToolSandbox, and ToolTalk.}
\label{fig:budget-recovered}
\end{figure}

The intensity levels correspond to the configurations in Table~\ref{tab:intensity-settings} of Supplement~\ref{app:overview}, and where both retrieval depth and step count vary, the comparison reflects their joint setting rather than either parameter in isolation. Fig.~\ref{fig:budget-recovered} addresses the separate question of resource availability by varying token and call limits while keeping method-specific settings intact.

\begin{figure}[t]
    \centering
    \begin{minipage}{\linewidth}
    \centering
    \subfigure[Cross-backbone rank agreement.]{\includegraphics[width=0.48\linewidth]{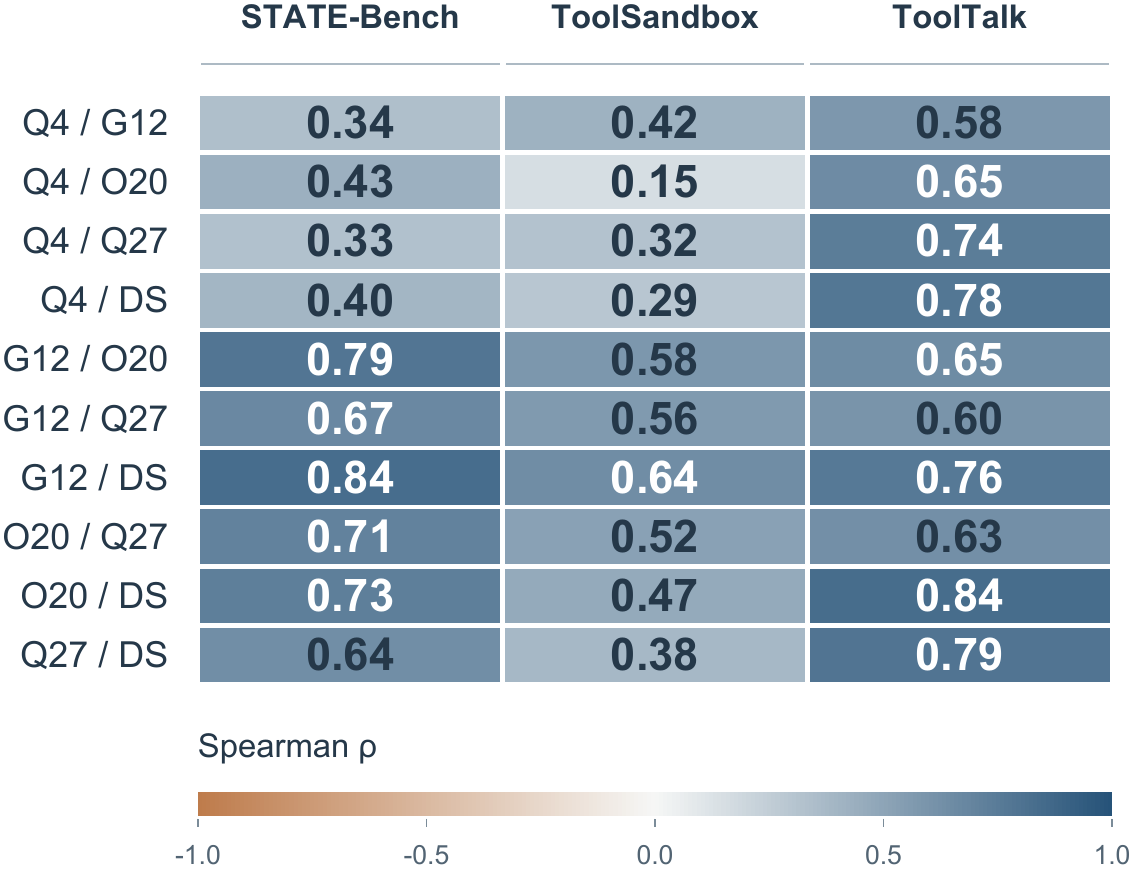}%
    \label{fig:dependencies-backbone}}%
    \hfill%
    \subfigure[Cross-arena rank agreement.]{\includegraphics[width=0.48\linewidth]{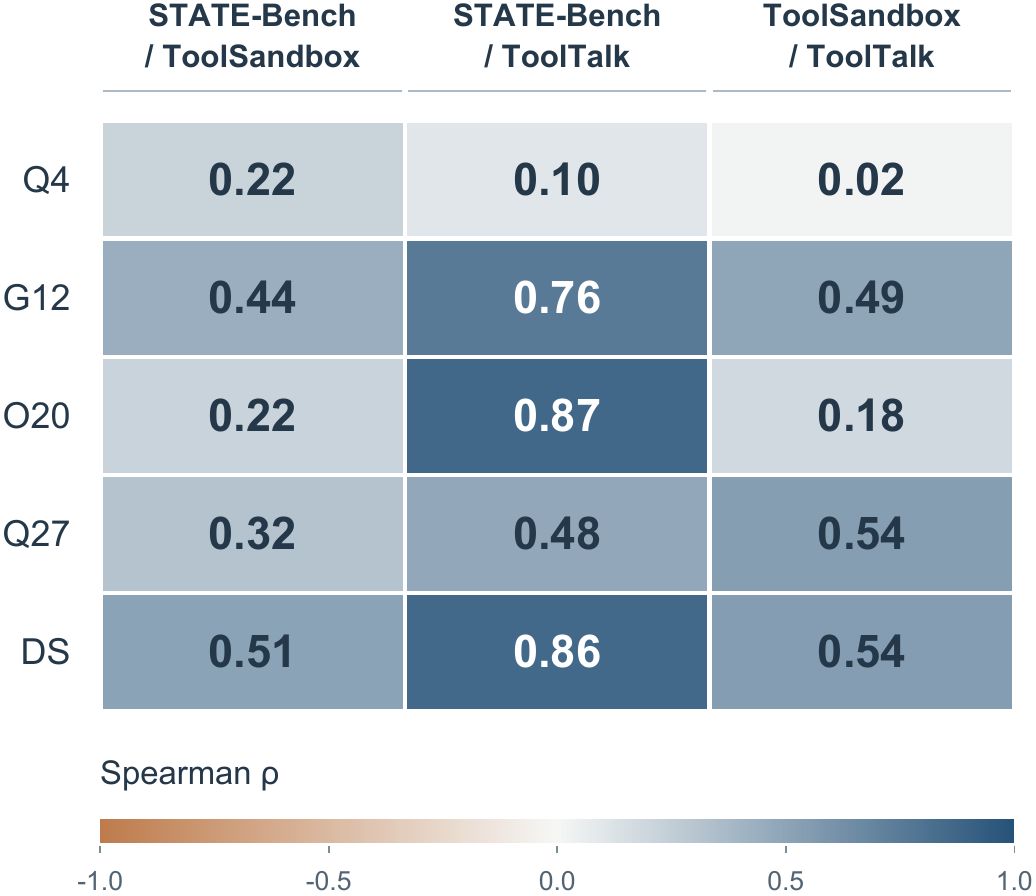}%
    \label{fig:dependencies-arena}}%
    \end{minipage}
    \caption{\textbf{Ranking agreement across five backbones.} Spearman correlations over the 23 methods, excluding \code{OPT0}, use unrounded Valid scores and average ranks for ties. (a) Backbone pairs within each arena. (b) Arena pairs within each backbone.}

    \Description{
        Two heatmaps summarize method-ranking agreement.
        The left panel compares ten backbone pairs across three
        arenas. The right panel compares three arena pairs across
        five backbones. Each cell contains a Spearman correlation
        computed over the same 23 intervention methods.
    }
    \label{fig:dependencies}
\end{figure}

\subsubsection{RQ4: Method rankings depend on deployment conditions}\label{sec:rq4}Sensitivity analyses test whether a ranking can guide deployment beyond the configuration in which it was measured. Fig.~\ref{fig:dependencies} shows that neither backbone nor arena changes preserve rankings uniformly. Across backbones within an arena, Spearman's $\rho$ ranges from 0.15 to 0.84, with Q4 and O20 agreeing more closely on ToolTalk than on ToolSandbox. Across arenas within a backbone, correlations range from 0.02 to 0.87, where Q4's near-zero ToolSandbox and ToolTalk agreement contrasts with O20's strong STATE-Bench and ToolTalk agreement. Rankings can therefore transfer across some settings while shifting markedly across others, so pipeline selection should be checked for the target arena and backbone pair. This pattern persists under Official scoring. Table~\ref{tab:metric-sensitivity} of Supplement~\ref{app:retrieval-archive} ranks all 23 interventions using the same main-table denominators: the top-scoring set matches Valid in 13 of 15 settings, while correlations still range from $-0.31$ to $0.86$. The two leader shifts occur on O20 and are detailed in Supplement~\ref{app:retrieval-archive}. Furthermore, on the shared scored-task subsets in
Table~\ref{tab:score-evidence}, 87 of the 115 cross-arena reversals
observed under either metric recur under both. 
Here, a reversal means that the score difference from the standard \code{OPT0} reference changes sign across two arenas for the same backbone, using the scored-task subsets in Table~\ref{tab:score-evidence}.
Score gaps like those in Table~\ref{tab:score-evidence} alter specific intervention choices, yet the broader task dependence also appears in native task scores.

The BRIGHT Earth Science comparison in Table~\ref{tab:bright-selected} of Supplement~\ref{app:retrieval-archive} and the resource-budget study in Fig.~\ref{fig:budget-recovered} show how execution choices change the size of a benefit. Query2doc increases nDCG@10 across all eight settings for each model, with paired changes from 0.047 to 0.156 for Q4 and from 0.057 to 0.269 for O20. RRR gains from larger budgets across all three arenas, with its O20 score rising from 3.0 to 33.3 on STATE-Bench while gains taper between the standard and larger budgets on ToolSandbox and ToolTalk. These findings place retriever choice, retrieval depth, and resource limits within the deployment decision, as benefit size varies with the execution setting even when its direction holds.

\begin{figure}[t]
\centering
\subfigure[]{\includegraphics[width=0.48\columnwidth]{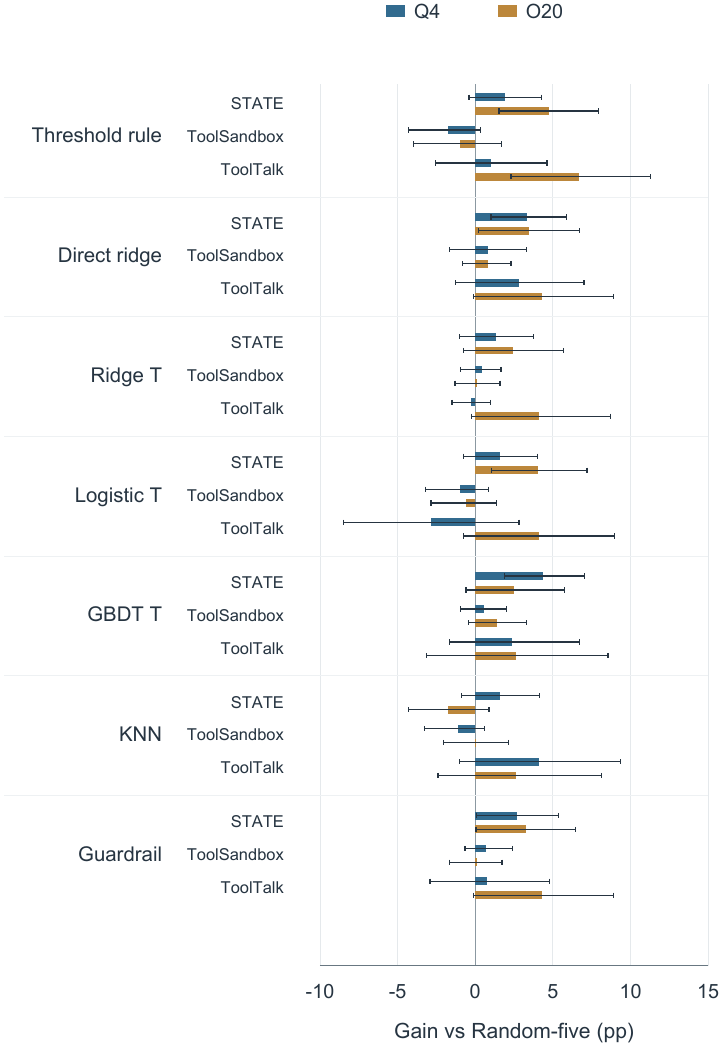}%
\label{fig:selector:gain}}%
\hfill%
\subfigure[]{\includegraphics[width=0.48\columnwidth]{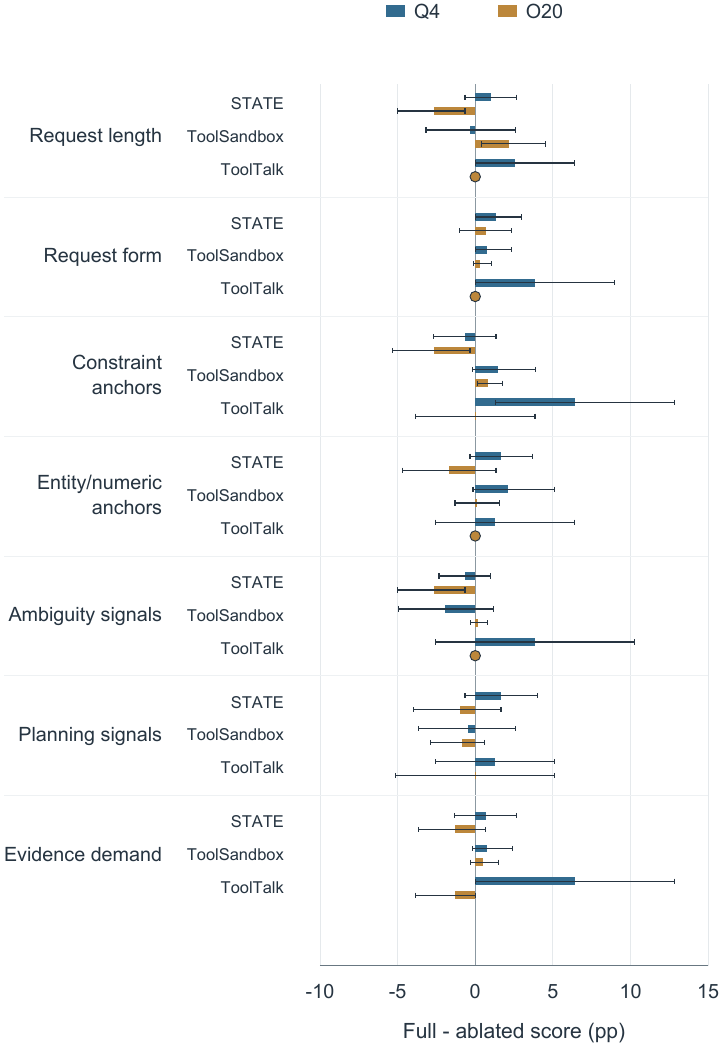}%
\label{fig:selector:features}}%
\caption{\textbf{Routing gains and feature contributions for Q4 and O20.} (a) Valid-score gain over Random-five at the same intervention rate. (b) Each bar compares Direct ridge using all seven feature blocks with
the same policy refitted and recalibrated without the indicated block.
Positive values mean that retaining the block yields a higher Valid score.
Differences are in percentage points. Whiskers show 95\% paired-bootstrap
confidence intervals for each comparison, with fitted policies held fixed.
A dot replaces the whisker when the two endpoints coincide.}
\Description{Two bar-chart panels show routing gains and seven feature-block
ablations. Each panel retains STATE-Bench, ToolSandbox,
ToolTalk, Q4, and O20 separately.}
\label{fig:selector}
\end{figure}

\subsubsection{RQ5: Routing gains depend on policy and input signals} 
We use the seven selectors to examine how pre-action information
supports choices among eligible pipelines and \texttt{OPT0}.
Three references answer different questions.
Rate-matched random selection controls intervention frequency
and tests whether a policy's action allocation improves on
uniform selection among the five pipelines.
A development-selected fixed action tests whether varying
the choice across inputs adds value over a constant policy.
The observed-outcome Oracle in Fig.~\ref{fig:motivation-headroom} gives a retrospective
limit within the recorded outcome table; its gap does not
establish that the available pre-action features can realize
that improvement.
These comparisons distinguish variation among
recorded outcomes from the utility achieved by fitted selectors.
Supplement.~\ref{app:routing-theory} provides an information-and-prediction
framework for interpreting this distinction.

\begin{figure}[t]
\centering
\includegraphics[width=0.9\columnwidth]{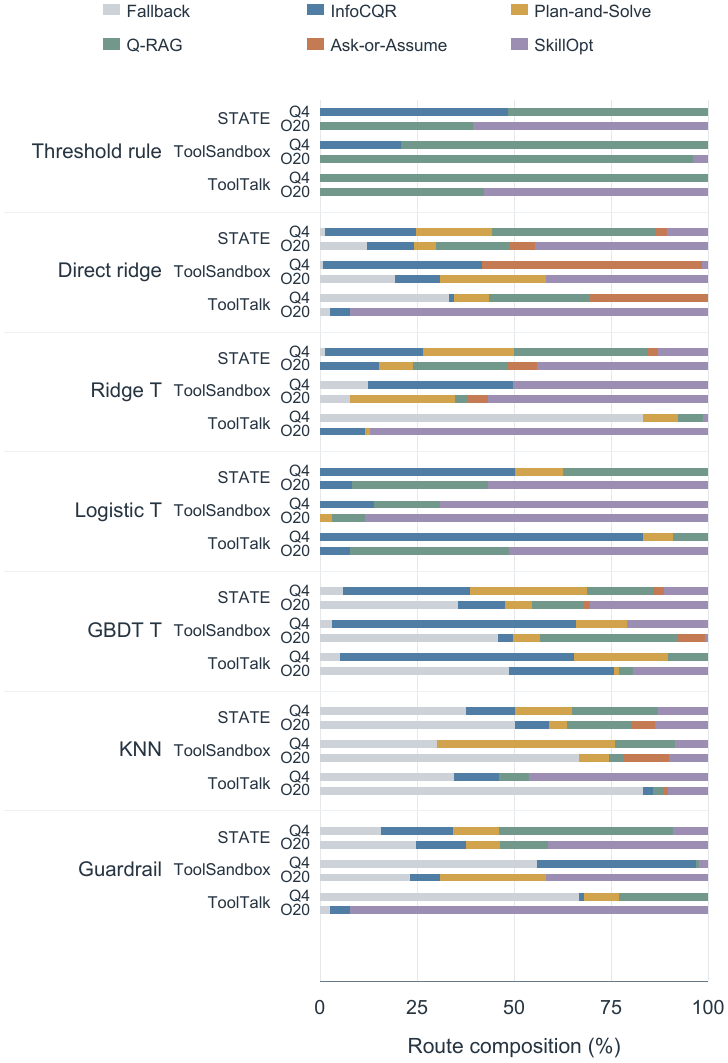}
\caption{\textbf{Action composition of the routing policies.}
Stacked bars show the five representative pipelines and \code{OPT0} shares for
the policies in Fig.~\ref{fig:selector}, fitted to $Y^T$ outcomes,
across Q4 and O20 evaluation cohorts.}
\Description{Six condition panels show the six-action route proportions
of all seven policies, separately for the three arenas and two backbones.}
\label{fig:selector:routes}
\end{figure}

Across all six arena and backbone pairs, GBDT T achieves positive point estimates over frequency-matched Random-five, ranging from $+0.58$ to $+4.36$ percentage points, abbreviated as pp. Policy ordering shifts with the arena as well: Threshold rule reaches $+6.67$ pp on ToolTalk under O20, while recording $-1.76$ pp on ToolSandbox under Q4 and $-1.00$ pp under O20.
The Fallback share in Fig.~\ref{fig:selector:routes} records how often the router selects \code{OPT0} before any intervention runs. On STATE-Bench, Guardrail selects \code{OPT0} more frequently than Direct ridge, showing lower absolute Valid scores even though its gains over matched random choices remain positive. 
Feature ablations in Fig.~\ref{fig:selector}(b) show which observations support these decisions. Request form is the most consistent feature block, yielding nonnegative point estimates across all six settings with a mean gain of 1.16 pp and a paired 95\% interval of $[0.34,2.08]$. Evidence demand and planning signals are less portable across settings; for example, evidence demand helps on ToolTalk under Q4 but shows the opposite sign under O20. 
The ablations favor retaining request-form signals in the
evaluated policy representation, while the contributions of
evidence-demand and planning signals vary across settings.
Pointwise intervals summarize individual differences, and small changes do not establish a fixed ranking among feature blocks. Section~\ref{app:selector-extension} evaluates benchmark-heldout transfer using the same 14-channel policy definitions and held-out arena refits.

\subsection{Benchmark-Heldout Evaluation}\label{app:selector-extension}
We test transfer by refitting policies for each held-out arena: STATE-Bench is evaluated after fitting on ToolSandbox, ToolSandbox is evaluated after fitting on STATE-Bench, and ToolTalk is evaluated after fitting on both. The 14 pre-action channels, seven policy definitions, and calibration grids remain unchanged. Normalization constants and policy parameters are refitted separately for each backbone and fold using its training groups, while thresholds are set on separate calibration groups.
Here, training refers to the policy-fitting partition. In each fold, eligible non-held-out STATE-Bench and ToolSandbox groups are divided into training and calibration sets under the same seed-2027 two-to-one rule. For each backbone, the test sets of the three held-out folds jointly
cover 657 distinct task groups: 450 from STATE-Bench, 129 from
ToolSandbox, and 78 from ToolTalk.
Table~\ref{tab:heldout-splits} reports the disjoint fitting, calibration, and held-out memberships. ToolTalk's evaluation set stays outside every fitting and calibration pool to provide a consistent cross-arena transfer test, and the STATE-Bench held-out fold evaluates all 450 groups, including the 150 development groups from the primary routing study.

Transfer shifts preferred routing behavior: Guardrail and Threshold rule attain the highest observed learned-policy scores on STATE-Bench for Q4 and O20, respectively. 
In the ToolSandbox offline task-group comparison, the aggregated
OPT0 reference scores highest for Q4, while Ridge T leads for
O20. Logistic T reaches the highest observed learned-policy
score on ToolTalk for both backbones.
Even with a common feature representation, preferred decision rules still vary across environments, so deployment requires checking both action selection and fallback behavior on the target task distribution. 
On O20 STATE-Bench, the fixed-action reference scores 23.6, compared with 20.9 for the highest-scoring learned policy, whereas on O20 ToolTalk Logistic T ties it at 51.3.
Beating frequency-matched random selection does not guarantee beating a development-selected fixed pipeline, so the two references compare routing against random action selection and a strong fixed pipeline, respectively.

\begin{table}[t]
\centering\small
\caption{Task-group counts for each benchmark-heldout fold, separately for Q4 and O20.}
\label{tab:heldout-splits}
\begin{tabular}{lrrr}
\toprule
Heldout arena & Training & Calibration & Test \\
\midrule
STATE-Bench & 86 & 43 & 450 \\
ToolSandbox & 300 & 150 & 129 \\
ToolTalk & 386 & 193 & 78 \\
\bottomrule
\end{tabular}
\end{table}

\begin{figure}[t]
\centering
\includegraphics[width=.97\columnwidth]{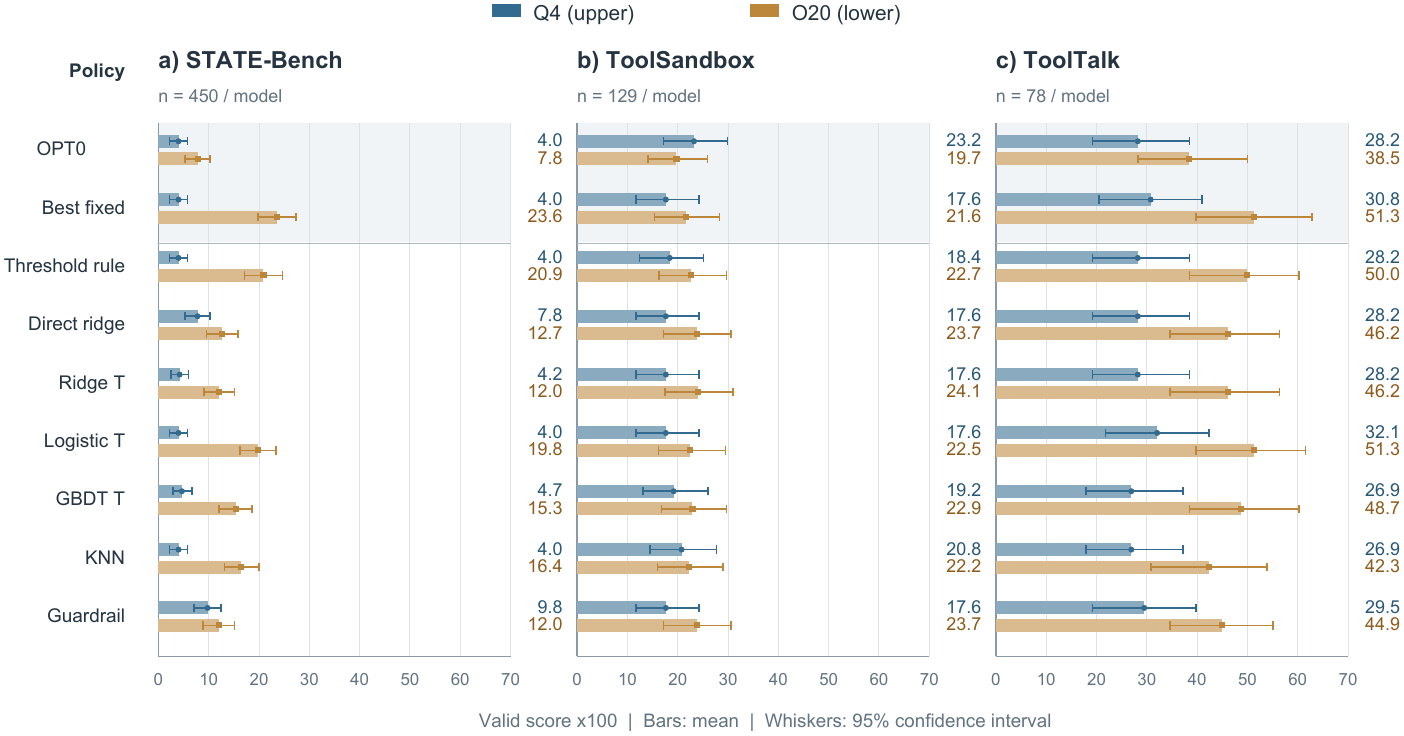}
\setlength{\abovecaptionskip}{4pt}
\caption{\textbf{Benchmark-heldout routing across three arenas.}
Policies are refitted on the development arenas, with ToolTalk always test-only.
Horizontal bars show Valid scores for Q4 (above) and O20 (below), alongside
\code{OPT0} and a training-selected fixed action. Whiskers show pointwise 95\%
task-group bootstrap intervals.}
\Description{Three grouped horizontal-bar panels show Q4 and O20 scores
for the seven policies and their two reference policies, with confidence intervals.}
\label{fig:retrospective-loso}
\end{figure}

\section{Discussion}\label{sec:disc}
\noindent\textbf{Conditional intervention makes non-intervention part of the decision.}
Keeping \code{OPT0} alongside active intervention candidates elevates leaving the input unchanged to an explicit decision evaluated under the exact same task criterion. Rather than assuming that every input benefits from preprocessing, a selector can leave the input unchanged whenever intervention is unnecessary or counterproductive.

\par\noindent\textbf{Intervention intensity and resource budgets have distinct effects.}
Separating intervention intensity from resource allocation clarifies two different configuration choices: intensity comparisons indicate how aggressively an operation should transform the input, whereas budget comparisons determine the token and call limits available to carry out that execution.

\noindent{\textbf{Fixed actions remain competitive.}
Our held-out results show that learned selectors do not
consistently outperform development-selected fixed actions.
Evaluations should therefore include both fixed and
rate-matched random baselines to assess the added value
of input-dependent selection.
}

\par\noindent\textbf{Execution profiles separate upstream integration from controlled comparison.}
Shared interfaces make upstream configuration choices explicit and reproducible across backends. Reporting profile-specific outcomes further distinguishes whether a method can be reliably integrated and executed from how well it performs under a specific deployment budget.

\par\noindent\textbf{Evidence strength and scope.}
Our analysis combines checks on scoring sensitivity,
execution repeatability, and task-sampling uncertainty.
Official and Valid scoring retain the same leading-method
set in 13 of 15 settings and share 87 of the 115
cross-arena gain reversals observed under either metric.
Three executions of 30 prespecified tasks per
backbone--arena setting further assess the repeatability
of paired pipeline gains on those tasks.
These checks support conclusions about recurring
gain directions and cross-setting reversals within the
evaluated configurations.
The routing bootstrap intervals are pointwise and conditional
on the fitted policies and recorded outcomes; they quantify
task-sampling uncertainty without incorporating variability
from policy refitting or repeated execution.
Fine-grained rankings, local intensity optima, and small
routing gains therefore remain exploratory.
Further evaluation can examine policy-refitting variability
and prospective transfer to new task distributions.


\section{Conclusion}
\textsc{AgentInputLab} offers a unified toolkit for task-aware input preparation in LLM agents, supporting both aggregate pipeline evaluation and paired task-level analysis. Across locally served and hosted backbones, top-performing methods vary by arena, and cross-arena rankings can be nearly uncorrelated under both Official and Valid scoring. Fixed-task repeats distinguish consistent gain directions from execution-sensitive fluctuations, while task-specific experiments show that preferred methods also shift across retrieved evidence, answers, and summaries.
{
These results support three concrete evaluation practices:
report intervention completion alongside native task success,
calibrate transformation strength under an explicit execution
budget, and compare input-dependent policies with both
\texttt{OPT0} and a development-selected fixed action.
By linking task outcomes to execution records, \textsc{AgentInputLab} provides a reusable
basis for assessing input preparation under declared execution
profiles and identifying settings in which fixed choices
remain competitive.
}

\bibliographystyle{IEEEtran}
\bibliography{references}




\clearpage
\setcounter{section}{0}
\renewcommand{\thesection}{EC.\Alph{section}}
\renewcommand{\theHsection}{supp.\Alph{section}}
\twocolumn[\centering{\LARGE\bfseries Supplement for \textsc{AgentInputLab}\par}\vspace{1em}]
\suppressfloats[t]
\section{Interfaces and Controlled Evaluation}\label{app:overview}\label{app:result-accounting}\label{app:rq-protocols}\label{app:fair}\label{app:interface}\label{app:plugin-recipes}

Dataset plugins expose the visible state, restorable environment, task-group identity, capability mask, and official scorer, while method adapters convert the visible state into upstream inputs and return patches over declared fields. Model, retrieval, tool, metric, and selector plugins supply the remaining execution interfaces, keeping intervention logic decoupled from model serving and benchmark-specific state restoration. Reproduction manifests record method and dependency revisions alongside task and response identities.
Table~\ref{tab:mainbench} summarizes the registered interfaces in the public release. To balance compute cost and cross-method comparability, we prioritize seven evaluated task settings: three share five-family support, and four extend the comparison to task-specific outputs, with additional results on BRIGHT Earth Science, FiQA, MultiFieldQA-en, and QMSum reported in Supplement~\ref{app:auxiliary-four}.

\begin{table*}[t]
\setlength{\abovecaptionskip}{-0.05cm}
\setlength{\belowcaptionskip}{-0.1cm}
\caption{The 31 registered dataset interfaces in public package
v0.1.0a14, with related tasks grouped and evaluated settings first.
Family codes describe registered support.
For the three main arenas, Table~\ref{tab:methodresults} reports
task-completion success for STATE-Bench, official similarity for
ToolSandbox, and task success for ToolTalk, all scaled to 0--100.
For the four additional task settings,
Table~\ref{tab:auxiliary-four} evaluates R/D/A methods on retrieval
and R/D/A/U methods on long-document tasks.
R, D, A, C, and U denote reformulation, decomposition, acquisition,
clarification, and curation, respectively.
``All'' denotes all five families and the identity action,
subject to instance eligibility.}
\label{tab:mainbench}
\centering
\scriptsize
\setlength{\tabcolsep}{3.5pt}
\renewcommand{\arraystretch}{1.12}
\resizebox{\textwidth}{!}{
\begin{tabular}{@{}lllll@{}}
\toprule
Interface & Task form & Registered family support & Official evaluation & Study role \\
\midrule
STATE-Bench \citep{microsoft2026statebench}
& Enterprise tool workflows
& All
& Task-completion success; UX and cost diagnostics
& Main: workflow and end state \\

ToolSandbox \citep{lu2025toolsandbox}
& Stateful conversational tool use
& All
& Official similarity (milestone and minefield evaluation)
& Main: tools and ambiguity \\

ToolTalk
& Stateful conversational tool use
& All
& Official task-success rate
& Main: conversational tool execution \\

BRIGHT Earth Science \citep{su2025bright}
& Reasoning-intensive scientific retrieval
& R/A/D
& nDCG and recall
& Additional results: scientific retrieval \\

FiQA (BEIR, development) \citep{thakur2021beir}
& Financial question retrieval
& R/A/D 
& nDCG@10
& Additional results: financial retrieval \\

MultiFieldQA-en (LongBench) \citep{bai2024longbench}
& Multi-domain long-document QA
& R/D/A/U
& Answer F1
& Additional results: document QA \\

QMSum (LongBench) \citep{bai2024longbench}
& Query-focused meeting summarization
& R/D/A/U
& ROUGE-L
& Additional results: summarization \\

\midrule
ArguAna \citep{thakur2021beir}
& Counterargument retrieval
& R/A/U
& nDCG and recall
& Available interface \\

BRIGHT Economics \citep{su2025bright}
& Reasoning-intensive economic retrieval
& R/A/U
& nDCG and recall
& Available interface \\

MuSiQue \citep{trivedi2022musique}
& Compositional multi-hop QA
& D/A/U
& Answer and support F1
& Available interface \\

2WikiMultiHopQA \citep{ho2020twowiki}
& Multi-hop QA with evidence triples
& D/A/U
& Answer, evidence, and joint F1
& Available interface \\

HotpotQA \citep{yang2018hotpotqa}
& Multi-hop QA with distractors
& D/A/U
& Answer, support, and joint F1
& Available interface \\

QASPER \citep{dasigi2021qasper}
& Long-document scientific QA
& A/U
& Answer and evidence F1
& Available interface \\
CQADupStack, Quora, DBPedia-Entity, FEVER & Retrieval (BEIR) & R/A & nDCG@10, recall & Available interface \\
BRIGHT Robotics, StackOverflow & Reasoning-intensive retrieval & R/D/A & nDCG@10, recall & Available interface \\
TREC DL 2019, 2020 & Passage retrieval & R/A & nDCG, MAP & Available interface \\
ClariQ & Clarification & R/C & Clarification utility & Available interface \\
LongBench QASPER & Document QA & D/A/U & Answer F1 & Available interface \\
LongBench GovReport & Summarization & U & ROUGE & Available interface \\
LongBench PassageRetrieval-en & Passage identification & A/U & Passage-ID accuracy & Available interface \\
LongBench RepoBench-P & Code completion & D/U & Exact match, code metric & Available interface \\
BFCL v3 Simple Python & Function calling & R/D & AST accuracy & Available interface \\
$\tau^3$ Banking Knowledge / Retail & Interactive tool use & R/D/C/U & Reward & Available interface \\
AppWorld-UL & Environment tasks & R/D; A/C/U conditional & Task success & Available interface \\
ALE External Transfer & Transfer tasks & R/D/A/U & Task score & Available interface \\
\bottomrule
\end{tabular}
}
\end{table*}

\noindent\textbf{Controlled comparisons and resource accounting.}
Algorithm~\ref{alg:unified-interface} defines the paired execution transaction, while \code{OPT0} uses the tool-feedback policy in Section~\ref{sec:exp:setup}. The intervention record $\ell_{i,j}$ stores execution status,
validation outcomes, and resource usage for scoring and accounting.
For \code{OPT0}, the identity record contains no intervention-stage
failure and zero intervention cost. Each intervention pipeline and its paired \code{OPT0} run
start from the same restored state under the controlled
settings listed in Table~\ref{tab:fairness}. We trace all 23 interventions to their upstream revisions and arena entry points, recording each source revision alongside its entry point
and file hashes in the experiment manifests. Table~\ref{tab:comparison-evidence} of Supplement~\ref{app:component} summarizes component verifications for the five representatives.
Resource accounting spans both the optimizer and downstream agent, deducting optimizer tokens and calls before downstream execution begins. The standard intervention and \code{OPT0} profile permits up to 4,096 output tokens and eight model calls, while stage-specific allowances and controlled adapter settings
are fixed in the experiment profiles under this cumulative budget rather than requiring identical internal calls. Simulator costs are recorded separately from the evaluated agent.

\begin{algorithm}[t]
\caption{Unified adapter and fair paired execution}
\label{alg:unified-interface}
\small
\begin{algorithmic}[1]
\Require Task $i$, method adapter $A_m$, policies $\pi_{\theta,0}$ and
$\pi_{\theta,m}$, comparison profile $\theta$

\State $(x_i,z_i,\textit{sourceRecord}_i)
  \gets \textsc{RestoreAndHash}(i)$
\State \textsc{ValidateEligibility}$(m,\mathcal{E}(x_i))$
\State $(z_{i,0},z_{i,m})
  \gets \textsc{IndependentRestore}(z_i)$
\State $\Delta_{i,0}\gets\varnothing$
\State $\ell_{i,0}\gets\textsc{IdentityRecord}()$

\State $\textit{upstreamInput}
  \gets\textsc{ToUpstreamInput}(A_m,x_i,\theta)$
\State $(\textit{upstreamOutput},\ell_{i,m})
  \gets\textsc{RunMethodProfile}$
\Statex \hspace{\algorithmicindent}
  $(A_m,\textit{upstreamInput},\theta)$
\State $\Delta_{i,m}
  \gets\textsc{ToTypedPatch}(\textit{upstreamOutput})$
\State $\ell_{i,m}\gets\textsc{ValidatePatch}
  (\Delta_{i,m},x_i,\theta,\ell_{i,m})$

\For{$j\in\{0,m\}$}
  \State $\tilde{x}_{i,j}
    \gets\textsc{Apply}(x_i,\Delta_{i,j})$
  \State $\tau_{i,j}
    \gets\pi_{\theta,j}(\tilde{x}_{i,j},z_{i,j})$
  \State $(Y^O_{i,j},G_{i,j},Y^T_{i,j})
    \gets\textsc{OfficialAndValidScore}$
  \Statex \hspace{\algorithmicindent}
    $(\tau_{i,j},\ell_{i,j})$
  \State $\textit{telemetry}_{i,j}
    \gets\textsc{CollectTelemetry}(\tau_{i,j},\ell_{i,j})$
\EndFor

\State \Return paired outcomes, telemetry, failures, and $\textit{sourceRecord}_i$
\end{algorithmic}
\end{algorithm}

\begin{table}[t]
\setlength{\abovecaptionskip}{-0.05cm}
\setlength{\belowcaptionskip}{-0.1cm}
\caption{Shared settings in controlled comparisons. Factors examined in the intensity and dependency studies vary
according to their specified settings.}
\label{tab:fairness}
\small
\begin{tabularx}{\columnwidth}{lY}
\toprule
Component & Matched setting \\
\midrule
Agent state & Same task checkpoint and visible history within each intervention--\code{OPT0} pair \\
LLM & Same checkpoint/API model and decoding parameters \\
Retrieval & Same corpus, chunks, index, retriever, reranker, $k$ \\
Knowledge & Same snapshot and permission boundary \\
Budget & Declared pipeline and reference token/call limits \\
Tools & Same schemas, simulator state, and side-effect policy \\
Scoring & Same official scorer and declared Valid-score rule \\
Randomness & Paired task order and prespecified seeds \\
\bottomrule
\end{tabularx}
\end{table}

\noindent\textbf{Task pairing and input groups.}
Paired analyses on Q4 and O20 evaluate the five representative pipelines alongside \code{OPT0} on the exact pre-action states associated with each pipeline, yielding 2,535 method-task differences from the 507 arena-specific groups per model. Fig.~\ref{fig:alignment-supplement} stratifies these outcomes by method-specific pre-action demand. Under this stratification, Matched indicates a reformulation opportunity for InfoCQR; multi-step execution dependencies for Plan-and-Solve; coverage by the frozen retrieval source for Q-RAG, including operation-level API documentation on ToolTalk; missing user intent or required user input for Ask or Assume?; and an explicit evidence-facing answer request for SkillOpt, such as lookup, verification, comparison, grounded recommendation, or numerical derivation. All remaining inputs are labeled Off-target, with identical labels applied to both backbones. 
We construct exploratory input-demand strata from pre-action requests, tool definitions, and frozen retrieval resources, combining deterministic rules with assistant-assisted case-level semantic judgments. The finalized labels are held fixed across backbones and score aggregation.
For each backbone, method, and subgroup, we first compute the mean gain over \code{OPT0} within each arena and then combine the three arena means with equal weights. Confidence intervals are estimated from 2,000 bootstrap draws of input-matched task clusters within each arena. Each draw applies the same cluster weights across methods and backbones while sampling arenas independently, contributing to an interval only when all three subgroup averages are defined. All intervals condition on the assigned subgroup labels.

\noindent\textbf{Method-Intensity Settings.}
Table~\ref{tab:intensity-settings} defines the parameters for Low, Moderate, and High across all three arenas using identical definitions. These settings adjust each method's intervention behavior, while resource-budget experiments vary the common execution limit.

\begin{table}[t]
\centering\small
\caption{Method-intensity settings. Parameters differ because the methods
control different operations: generation, planning/retrieval, clarification,
or compression. Stage allowances remain subject to the shared profile.}
\label{tab:intensity-settings}
\scriptsize\setlength{\tabcolsep}{3pt}
\begin{tabularx}{\columnwidth}{@{}lYccc@{}}
\toprule
Method & Parameter & Low & Moderate & High \\
\midrule
Query2doc & Optimizer output-token allowance & 96 & 192 & 384 \\
Decomp & Maximum steps / retrieval top-$k$ & 2 / 2 & 4 / 5 & 8 / 10 \\
IRCoT & Maximum steps / retrieval top-$k$ & 1 / 2 & 2 / 5 & 4 / 10 \\
UDCS & Clarification threshold / optimizer-token allowance & 0.8 / 200 & 0.5 / 400 & 0.2 / 800 \\
LLMLingua-2 & Retained-token ratio & 0.8 & 0.5 & 0.25 \\
\bottomrule
\end{tabularx}
\end{table}

\begin{figure}[t]
\centering
\includegraphics[width=0.98\columnwidth]{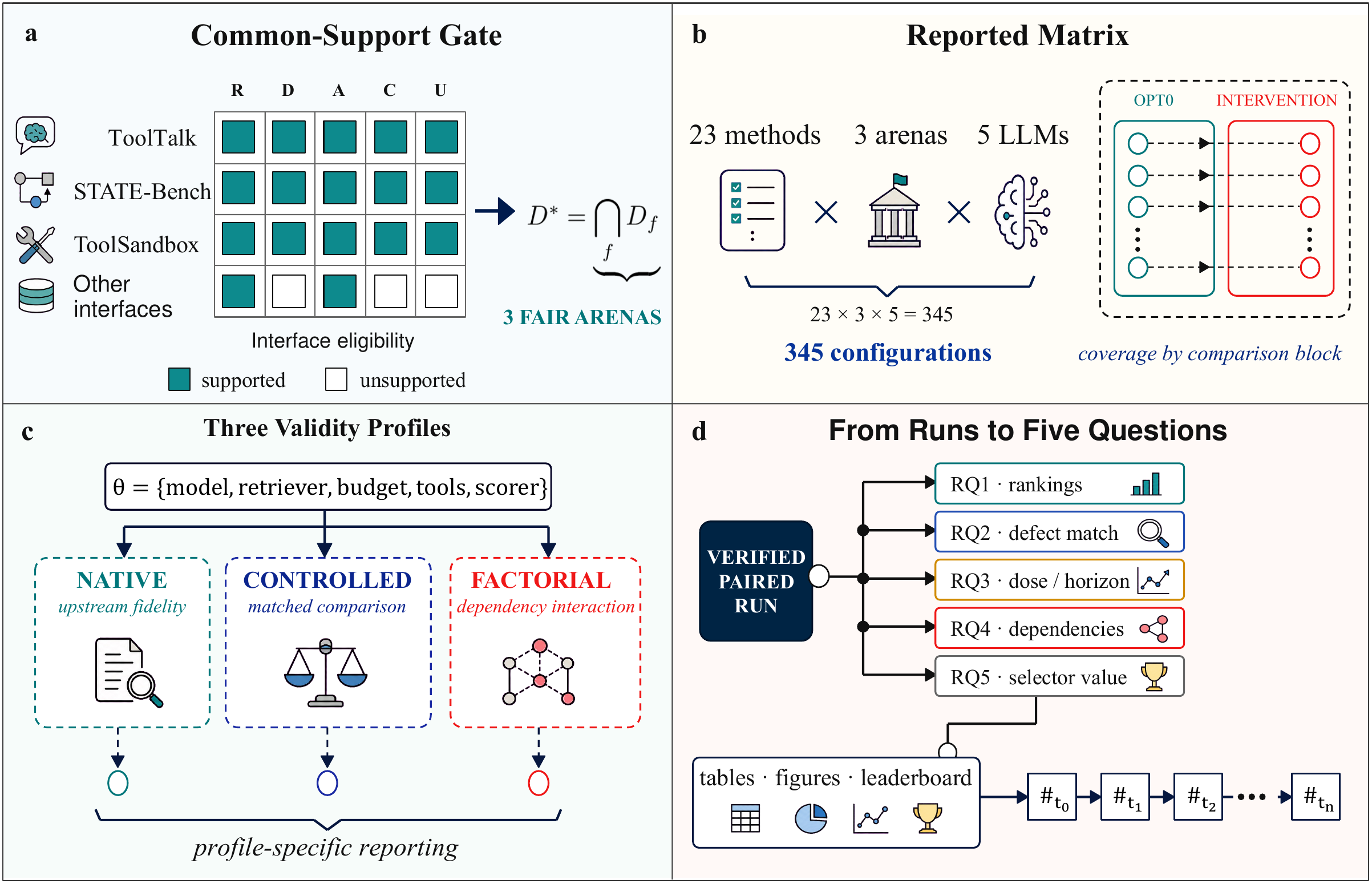}
\caption{\textbf{From compatibility to controlled evidence.}
The illustrated interfaces connect through task eligibility to
comparison profiles, paired execution, and result aggregation.}
\Description{Four panels show capability coverage, the comparison matrix,
execution profiles, and source records.}
\label{fig:protocol}
\end{figure}
\section{Routing Model Specifications}\label{app:selector}
\noindent\textbf{Input Representation and Calibration.} Tables~\ref{tab:figure6-features} and \ref{tab:figure6-policies} specify the input channels and fixed learners used in Fig.~\ref{fig:selector}. The primary routing study trains on 100 STATE-Bench development groups, calibrates on 50 disjoint groups, and evaluates on the remaining STATE-Bench groups plus all ToolSandbox and ToolTalk groups, allowing all three arenas to contribute outcomes while testing transfer on the latter two without within-arena fitting. The representation captures pre-action properties of each request: size, form, constraints, ambiguity, planning needs, and evidence demand. Constraint counts tally distinct requirement clauses, entity and numeric counts tally distinct visible anchors, and request-form signals count fragmentary, keyword-style, extreme-length, and noisy inputs. Ambiguity, planning, and evidence-demand signals apply fixed lexical rules to the request, supplying pre-action proxies without human defect labels. Density measures counts per 100 request characters with length bounded below by one, and entity and numeric-anchor density sums entity and number counts.

\begin{table}[t]
\centering\scriptsize
\caption{Seven blocks containing all 14 pre-action channels used in
Fig.~\ref{fig:selector}. Counts and lexical signals are observable proxies.}
\label{tab:figure6-features}
\begin{tabularx}{\columnwidth}{lY}
\toprule
Block & Channels \\
\midrule
Request length & Request character count \\
Request form & Question-mark count; request-form signal count \\
Constraint anchors & Constraint count; constraint density \\
Entity/numeric anchors & Entity count; number count; combined anchor density \\
Ambiguity signals & Ambiguity count; ambiguity density \\
Planning signals & Multi-step count; multi-step density \\
Evidence demand & Retrieval-demand count; retrieval-demand density \\
\bottomrule
\end{tabularx}
\end{table}

\begin{table}[t]
\centering\scriptsize
\caption{Fixed policies in Fig.~\ref{fig:selector}, using standard
supervised learners and separate action-outcome models
\citep{hastie2009,friedman2001,kunzel2019}.}
\label{tab:figure6-policies}
\begin{tabularx}{\columnwidth}{lY}
\toprule
Policy & Fixed numerical definition \\
\midrule
Threshold rule & Utility-maximizing one-split stump; at least 10 training rows per leaf \\
Direct ridge & Linear action-gain models; ridge $10^{-6}$ \\
Ridge T & Separate action-outcome regressions; ridge $10^{-3}$ \\
Logistic T & Separate action-outcome models; $\ell_2=0.001$, learning rate $0.2$, 600 iterations \\
GBDT T & NumPy squared-error boosting; 50 trees per action, depth two, learning rate $0.05$ \\
KNN & Five nearest neighbors; inverse squared-distance weights; exact matches averaged \\
Guardrail & Direct-ridge gains with fixed training-derived input-cost and execution-flag priors \\
\bottomrule
\end{tabularx}
\end{table}

All policies share the same 14 channels with backbone-specific parameters. Each raw channel $x_j$ is scaled via $x_j/(x_j+t_j)$, where $t_j$ is the median of positive training values or one if none exist, and these constants carry over unchanged to calibration and evaluation. Ordinary gain thresholds span $\{-0.05,0,0.05\}$ and margins span $\{0,0.02,0.05\}$. Guardrail optimizes adjusted gain $\widehat g-10^{-5}\overline C-0.1\overline H$, where $\overline C$ and $\overline H$ are action-specific training means of added input cost and execution flags, subject to cost limits at half or full maximum training cost, execution-flag limits $\{0.05,0.1,1\}$, and the same gain and margin grid. Guardrail also requires the base Direct-ridge rule to route first, with calibration ties resolved by higher routing coverage, fewer recorded flags, and a fixed grid order. 
The action space includes the five representative pipelines and
an aggregated \code{OPT0} reference. For each task group, the router chooses one of five representative
pipelines or \code{OPT0} (no input intervention). To score the
\code{OPT0} choice, we average the five method-paired \code{OPT0}
scores. Let $Y^T_{i,0\mid m}$ be the Valid score of the \code{OPT0}
run started from the same initial state as method $m$ on task group $i$.
The score assigned to \code{OPT0} is
\[
Y^T_{i,s}=\frac{1}{5}\sum_{m=1}^{5}Y^T_{i,0\mid m}.
\]
When all five methods share one reference run, this is simply its
score. On ToolSandbox, the references use method-specific initial
states because message and reminder timestamps can differ across
the archived environments, as described in Supplement~\ref{app:component}.

Each intervention entry retains the score from its corresponding
paired run. These task-group outcome tables are used for policy
fitting, calibration, and evaluation. For policy $\pi$ with intervention rate $r_\pi$, the matched Random-five baseline has expected score
\[
\bar Y^{\mathrm{rand}}_{\pi}
=(1-r_{\pi})\bar Y^T_s+
\frac{r_{\pi}}{5}
\sum_{a\in\{R,D,A,C,U\}}\bar Y^T_a.
\]
where $\bar Y^T_s$ is the mean of $Y^T_{i,s}$ over the evaluated
task groups, and $R,D,A,C,U$ denote the five representative
pipelines.
For feature-block comparisons, Direct ridge is refitted and recalibrated after dropping the full block, and 5,000 paired bootstrap draws condition on the frozen policies and observed cohort routing rates. In Fig.~\ref{fig:motivation}(b), SkillOpt had the highest mean Valid
score among the five representative pipelines on the STATE-Bench
training groups pooled across Q4 and O20. We therefore use SkillOpt
as the fixed baseline for both backbones in all three test arenas.
The Oracle instead takes the highest recorded Valid score for each
task, including \code{OPT0} among its choices.
In Fig.~\ref{fig:retrospective-loso}, the fixed baseline is chosen
anew for each backbone and held-out arena. We select the action
with the highest mean Valid score on that fold's training groups
and use it for every task in the held-out arena.

\section{Component Checks and Repeatability}\label{app:component}
For each of the five representative methods, we checked the key step listed in Table~\ref{tab:comparison-evidence} on 30 tasks per backbone--arena pair. With two backbones (Q4 and O20) and three arenas (STATE-Bench, ToolSandbox, and ToolTalk), this gave 180 runs per method and 900 in total. For Q-RAG, replaying the released selector with frozen weights on the same ordered candidates reproduced all 180 ranked document lists exactly. Of the 55 runs in which the intervention failed, 11 still earned positive Official scores. In one ToolTalk run with O20, for example, InfoCQR failed to produce a rewrite, but the agent completed the task using the original request. The Official score was 1 because the task succeeded; the Valid score was 0 because the rewrite did not complete.

\begin{table}[t]
\centering\scriptsize
\caption{Component-level checks for the five representatives. Each method has 180 cases: 30 per backbone--arena setting. Of the 51 Ask or Assume? outputs rejected (47 for structure or field values, four for JSON syntax).}
\label{tab:comparison-evidence}
\begin{tabularx}{\columnwidth}{@{}lYrr@{}}
\toprule
Method & Checked component & \shortstack{Matching\\traces} & \shortstack{Method\\failures} \\
\midrule
InfoCQR & Released rewrite call and returned query & 176/180 & 4 \\
Plan-and-Solve & Upstream planning trigger & 180/180 & 0 \\
Q-RAG & Selected documents and order & 180/180 & 0 \\
Ask or Assume? & Clarification/resumption transition & 129/180 & 51 \\
SkillOpt & Frozen skill block & 180/180 & 0 \\
\bottomrule
\end{tabularx}
\end{table}

\noindent\textbf{Repeatability on fixed tasks.}
Table~\ref{tab:repeat-evidence} asks whether a method still helps or hurts when the same tasks are run again. For each of the 30 method--backbone--arena combinations, we used the same 30 task groups in the original run and both repeats, retaining graded failures. On each task, the method and \code{OPT0} started from matching states. STATE-Bench and ToolTalk shared \code{OPT0} runs when those states were identical; ToolSandbox used a separate reference for each method because message and reminder timestamps differed across the archived environments. STATE-Bench repeats were independent executions because its agent requests did not pass a seed. For each run, we averaged the 30 paired score differences (method minus \code{OPT0}). The table reports the mean and standard deviation of these three run-level gains. A Wins entry of $3/3$ means that the average gain was positive in every run, not that the method won on every task.
Under both Official and Valid scoring, 14 of the 30 combinations had positive gains in all three runs, and two had negative gains in all three. On O20, all five methods improved on \code{OPT0} in every STATE-Bench run; for InfoCQR and Q-RAG, the pointwise 95\% intervals also stayed above zero each time. On Q4, InfoCQR illustrates two different kinds of reversal. It outperformed \code{OPT0} on STATE-Bench and underperformed it on ToolTalk in all three runs: changing the arena reversed which pipeline was preferable. On ToolSandbox, by contrast, InfoCQR's gain was negative in the original run and positive in both repeats. Here the direction changed even though the arena and tasks stayed the same.

\begin{table}[t]
\centering\scriptsize
\setlength{\tabcolsep}{1pt}
\caption{Persistence of paired gains across the original execution and two repeats. Positive values favor the method over its paired \code{OPT0}.}
\label{tab:repeat-evidence}
\begin{tabular*}{\columnwidth}{@{\extracolsep{\fill}}llrrrr@{}}
\toprule
& & \multicolumn{2}{c}{Gain (pp), mean $\pm$ SD} & \multicolumn{2}{c}{Wins (of 3)} \\
\cmidrule(lr){3-4}\cmidrule(l){5-6}
Setting & Method & Official & Valid & Official & Valid \\
\midrule
Q4--S & InfoCQR & $+5.56 \pm 1.92$ & $+5.56 \pm 1.92$ & 3 / 3 & 3 / 3 \\
 & Plan-and-Solve & $+5.56 \pm 1.92$ & $+5.56 \pm 1.92$ & 3 / 3 & 3 / 3 \\
 & Q-RAG & $+6.67 \pm 3.33$ & $+6.67 \pm 3.33$ & 3 / 3 & 3 / 3 \\
 & Ask or Assume? & $+7.78 \pm 6.94$ & $+4.44 \pm 5.09$ & 2 / 3 & 2 / 3 \\
 & SkillOpt & $+10.00 \pm 6.67$ & $+10.00 \pm 6.67$ & 3 / 3 & 3 / 3 \\
\addlinespace[3pt]
Q4--X & InfoCQR & $+0.34 \pm 3.83$ & $+0.34 \pm 3.83$ & 2 / 3 & 2 / 3 \\
 & Plan-and-Solve & $+0.15 \pm 3.18$ & $+0.15 \pm 3.18$ & 2 / 3 & 2 / 3 \\
 & Q-RAG & $+0.73 \pm 3.69$ & $+0.73 \pm 3.69$ & 2 / 3 & 2 / 3 \\
 & Ask or Assume? & $+1.63 \pm 4.34$ & $-0.71 \pm 8.40$ & 2 / 3 & 2 / 3 \\
 & SkillOpt & $-0.57 \pm 2.20$ & $-0.57 \pm 2.20$ & 1 / 3 & 1 / 3 \\
\addlinespace[3pt]
Q4--T & InfoCQR & $-6.67 \pm 0.00$ & $-6.67 \pm 0.00$ & 0 / 3 & 0 / 3 \\
 & Plan-and-Solve & $+6.67 \pm 0.00$ & $+6.67 \pm 0.00$ & 3 / 3 & 3 / 3 \\
 & Q-RAG & $-3.33 \pm 0.00$ & $-3.33 \pm 0.00$ & 0 / 3 & 0 / 3 \\
 & Ask or Assume? & $0.00 \pm 0.00$ & $-3.33 \pm 5.77$ & 0 / 3 & 0 / 3 \\
 & SkillOpt & $0.00 \pm 0.00$ & $0.00 \pm 0.00$ & 0 / 3 & 0 / 3 \\
\addlinespace[3pt]
O20--S & InfoCQR & $+22.22 \pm 6.94$ & $+22.22 \pm 6.94$ & 3 / 3 & 3 / 3 \\
 & Plan-and-Solve & $+16.67 \pm 3.33$ & $+16.67 \pm 3.33$ & 3 / 3 & 3 / 3 \\
 & Q-RAG & $+20.00 \pm 6.67$ & $+20.00 \pm 6.67$ & 3 / 3 & 3 / 3 \\
 & Ask or Assume? & $+13.33 \pm 3.33$ & $+13.33 \pm 3.33$ & 3 / 3 & 3 / 3 \\
 & SkillOpt & $+18.89 \pm 3.85$ & $+18.89 \pm 3.85$ & 3 / 3 & 3 / 3 \\
\addlinespace[3pt]
O20--X & InfoCQR & $+0.50 \pm 0.98$ & $+0.02 \pm 1.09$ & 2 / 3 & 2 / 3 \\
 & Plan-and-Solve & $+2.21 \pm 0.09$ & $+2.21 \pm 0.09$ & 3 / 3 & 3 / 3 \\
 & Q-RAG & $+1.18 \pm 2.17$ & $+1.18 \pm 2.17$ & 2 / 3 & 2 / 3 \\
 & Ask or Assume? & $+1.74 \pm 1.25$ & $+1.53 \pm 1.60$ & 3 / 3 & 2 / 3 \\
 & SkillOpt & $+0.64 \pm 0.90$ & $+0.64 \pm 0.90$ & 2 / 3 & 2 / 3 \\
\addlinespace[3pt]
O20--T & InfoCQR & $+4.44 \pm 7.70$ & $0.00 \pm 8.82$ & 1 / 3 & 1 / 3 \\
 & Plan-and-Solve & $+8.89 \pm 5.09$ & $+8.89 \pm 5.09$ & 3 / 3 & 3 / 3 \\
 & Q-RAG & $+8.89 \pm 1.92$ & $+8.89 \pm 1.92$ & 3 / 3 & 3 / 3 \\
 & Ask or Assume? & $-1.11 \pm 8.39$ & $-2.22 \pm 6.94$ & 1 / 3 & 1 / 3 \\
 & SkillOpt & $+10.00 \pm 5.77$ & $+10.00 \pm 5.77$ & 3 / 3 & 3 / 3 \\
\addlinespace[3pt]
\bottomrule
\end{tabular*}
\par\vspace{3pt}
\begin{minipage}{\columnwidth}\scriptsize
S, X, and T denote STATE-Bench, ToolSandbox, and ToolTalk. Every row uses all 30 prespecified task pairs in all three executions. SD is the sample standard deviation of the three gains, not a confidence interval. The last two columns count executions with a strictly positive gain; zero is a tie.
\end{minipage}
\end{table}

\section{Scoring and Configuration Sensitivity}
\label{app:retrieval-archive}\label{app:budget-all-models}\label{app:tooltalk-budget}
A method can outperform \code{OPT0} in one arena yet fall behind it
in another. Does this pattern depend on how we score the method?
Table~\ref{tab:score-evidence} compares Official and Valid on the
same scored tasks. The distinction matters when a method fails but
the downstream agent still earns credit: all 1,024 cases with a
positive Official score and a zero Valid score had a recorded method
failure. Even so, most reversals appear under either score. Holding
the backbone fixed, we found 87 comparisons in which a method went
from above \code{OPT0} in one arena to below it in another under
both scoring rules. Another 22 showed this reversal only under
Valid, and six only under Official.

Within a given arena, changing the score rarely changed which method
ranked first. We ranked all 23 interventions using the unrounded
main-table scores and kept ties. These averages include every planned
task: confirmed failures without a native grade contribute zero to
the average, while their individual grades remain missing. Official
and Valid gave the same winners in 13 of 15 backbone--arena settings.
Both exceptions were on O20. On STATE-Bench, Adaptive-RAG led under
Official and Q-RAG under Valid; on ToolSandbox, Decomp and RECOMP
led, respectively. Agreement across arenas was less consistent.
In Table~\ref{tab:metric-sensitivity}, the O20 rankings for STATE-Bench
and ToolSandbox were negatively correlated under Official but positively
correlated under Valid. By contrast, the DeepSeek-V4.1-flash rankings
for STATE-Bench and ToolTalk agreed strongly under both scores.
\begin{table}[t]
\centering\scriptsize
\caption{Cross-arena rank correlations for all 23 interventions using
the unrounded main-table aggregates and the same failure accounting.
S, X, and T denote STATE-Bench, ToolSandbox, and ToolTalk.}
\label{tab:metric-sensitivity}
\begin{tabular}{@{}llrr@{}}
\toprule
Model & Arena pair & Official $\rho$ & Valid $\rho$ \\
\midrule
Q4 & S--X & -0.20 & 0.22 \\
Q4 & S--T & 0.02 & 0.10 \\
Q4 & X--T & -0.23 & 0.02 \\
G12 & S--X & 0.28 & 0.44 \\
G12 & S--T & 0.75 & 0.76 \\
G12 & X--T & 0.34 & 0.49 \\
O20 & S--X & -0.31 & 0.22 \\
O20 & S--T & 0.73 & 0.87 \\
O20 & X--T & -0.06 & 0.18 \\
Q27 & S--X & -0.13 & 0.32 \\
Q27 & S--T & 0.34 & 0.48 \\
Q27 & X--T & 0.32 & 0.54 \\
DS & S--X & 0.44 & 0.51 \\
DS & S--T & 0.86 & 0.86 \\
DS & X--T & 0.49 & 0.54 \\
\bottomrule
\end{tabular}
\end{table}

\noindent\textbf{Retriever settings.}
The BRIGHT Earth Science comparison evaluates 116 identical query IDs per setting with random seed 17. Four retrievers, namely BM25, E5, BGE-M3, and Qwen3-Embedding, are tested at top-$k$ depths of 5 and 20, yielding eight settings per model. For Query2doc generation, Q4 uses a 256-output-token budget with thinking disabled alongside E5-large-v2, whereas O20 uses a 4,096-token budget with low reasoning alongside E5-base-v2. Raw and expanded queries share the exact retrieval configuration within each model-setting pair, and Table~\ref{tab:bright-selected} reports native 0 to 1 metrics. Query2doc improves nDCG@10 across every evaluated retrieval setting, yet its gain is substantially larger with BGE-M3 than with E5, and increasing retrieval depth further alters recall and resulting ranking quality. While the consistent positive direction supports query expansion in these settings, the wide variation in magnitude demonstrates that practical utility must be assessed jointly with the deployed retriever rather than inferred from an isolated reformulation score.

\begin{table}[t]
\centering\small
\setlength{\tabcolsep}{5pt}
\caption{BRIGHT Earth Science retrieval results. Raw is the original query
and Q2D is Query2doc.}
\label{tab:bright-selected}
\begin{tabular}{lrrrrr}
\toprule
& & \multicolumn{2}{c}{nDCG@10} & \multicolumn{2}{c}{Recall@$k$} \\
Retriever & $k$ & Raw & Q2D & Raw & Q2D \\
\midrule
\multicolumn{6}{l}{\textbf{Q4}: 8 settings $\times$ 116 queries = 928 paired comparisons} \\
BM25 & 5 & 0.105 & 0.165 & 0.095 & 0.153 \\
BM25 & 20 & 0.127 & 0.195 & 0.191 & 0.310 \\
E5-large-v2 & 5 & 0.202 & 0.257 & 0.211 & 0.244 \\
E5-large-v2 & 20 & 0.218 & 0.281 & 0.293 & 0.332 \\
BGE-M3 & 5 & 0.131 & 0.277 & 0.155 & 0.292 \\
BGE-M3 & 20 & 0.154 & 0.310 & 0.261 & 0.437 \\
Qwen3-Embedding & 5 & 0.225 & 0.285 & 0.246 & 0.298 \\
Qwen3-Embedding & 20 & 0.273 & 0.320 & 0.399 & 0.445 \\
\midrule
\multicolumn{6}{l}{\textbf{O20}: 8 settings $\times$ 116 queries = 928 paired comparisons} \\
BM25 & 5 & 0.105 & 0.280 & 0.095 & 0.263 \\
BM25 & 20 & 0.127 & 0.313 & 0.191 & 0.433 \\
E5-base-v2 & 5 & 0.139 & 0.197 & 0.151 & 0.186 \\
E5-base-v2 & 20 & 0.156 & 0.226 & 0.227 & 0.278 \\
BGE-M3 & 5 & 0.131 & 0.378 & 0.155 & 0.366 \\
BGE-M3 & 20 & 0.154 & 0.423 & 0.261 & 0.507 \\
Qwen3-Embedding & 5 & 0.228 & 0.333 & 0.246 & 0.349 \\
Qwen3-Embedding & 20 & 0.273 & 0.364 & 0.401 & 0.478 \\
\bottomrule
\end{tabular}
\end{table}

\noindent\textbf{Resource budgets.}
We evaluate twelve intervention pipelines and \code{OPT0} on Q4 and
O20 across all three arenas. At the standard ($1\times$) budget,
each intervention pipeline has up to 4,096 output tokens in total,
8 model calls, and 512 output tokens per call. The $0.5\times$ and
$2\times$ profiles halve or double all three limits without changing
the method settings. \code{OPT0} retains its tool-feedback loop at
each of its reference budgets.  Fig.~\ref{fig:budget-recovered} shows why the choice of budget matters
even after a pipeline has been selected. RRR benefits from larger
budgets in all three arenas on both backbones. For \code{OPT0} on O20,
however, a larger budget raises the score on STATE-Bench but lowered
it on ToolSandbox. Thus, even for the same pipeline and backbone,
a budget increase that helps in one arena can hurt in another.

\section{Additional Task Settings}\label{app:auxiliary-four}
We evaluate applicable methods across four static benchmarks: BRIGHT Earth Science with 116 queries\citep{su2025bright}, the FiQA development set through BEIR with 500 queries\citep{thakur2021beir}, and the LongBench MultiFieldQA-en and QMSum tasks with 150 and 200 queries, respectively\citep{bai2024longbench}. Retrieval tasks are evaluated with nDCG@10, document question answering with answer F1, and query-focused summarization with ROUGE-L. DeepSeek-V4.1-flash provides hosted generation while frozen local retrieval, compression, and routing components are retained; R2AG uses a local Q4 reader, and retrieval-only Q-RAG requires no hosted generation. Table~\ref{tab:auxiliary-four} evaluates 18 methods from four families:
11 methods from reformulation, decomposition, and acquisition on
each retrieval task, and all 18 methods, including seven curation
methods, on each long-document task.
The retrieval protocol scores document rankings, whereas the
long-document protocol also evaluates context curation before
answer or summary generation.
Methods requiring user feedback, conversation history, or an
environment tool-execution graph are excluded from these static
task evaluations. \code{OPT0} uses the same ReAct policy as in the main experiment
and searches through each task's native interface. All methods use
the official split for each task. The native scorer evaluates the
final ranking on retrieval tasks and the final answer on answer-based
tasks.
On retrieval tasks, Adaptive-RAG returns the ranking produced by
its selected branch. Plan-and-Solve, Plan-and-Budget, and R2AG
instead use a shared adapter to turn each method's output into a
search query. The retriever then returns the ranked documents.
This step leaves each method's core procedure unchanged and stays
within its original budget. The separate Q4 and O20 retriever
comparison is reported in Table~\ref{tab:bright-selected}
in Supplement~\ref{app:retrieval-archive}.

\begin{table}[t]
\centering\scriptsize
\setlength{\tabcolsep}{2.5pt}
\caption{Additional task results by family (Official / Valid),
on a 0--100 scale. Dashes mark methods outside the evaluated
task-specific profiles.
Reader control uses the same query-output interface without input
optimization; \code{OPT0} is the ReAct reference.}
\label{tab:auxiliary-four}
\begin{tabular}{lrrrr}
\toprule
Family / Method & BRIGHT ES & MultiFieldQA-en & QMSum & FiQA-dev \\
 & nDCG@10 & F1 & ROUGE-L & nDCG@10 \\
\midrule
\code{OPT0} & 17.8 / 17.8 & 54.5 / 54.5 & 22.7 / 22.7 & 27.1 / 27.1 \\
Reader control & 17.0 / 17.0 & -- & -- & 22.9 / 22.9 \\
\midrule
\multicolumn{5}{l}{\textit{R: Reformulation}} \\
Query2doc & 21.6 / 21.6 & 59.5 / 59.5 & 23.4 / 23.4 & 31.5 / 31.5 \\
RRR & 11.7 / 11.7 & 57.5 / 57.5 & 24.4 / 24.4 & 20.0 / 20.0 \\
ReFormeR & 15.0 / 15.0 & 57.5 / 57.5 & 22.9 / 22.9 & 21.8 / 21.8 \\
\midrule
\multicolumn{5}{l}{\textit{D: Decomposition}} \\
Plan-and-Solve & 17.1 / 17.1 & 57.4 / 57.4 & 23.5 / 23.5 & 24.0 / 24.0 \\
Decomp & 6.6 / 6.6 & 51.7 / 51.7 & 23.7 / 23.7 & 16.0 / 16.0 \\
Self-Ask & 1.6 / 1.6 & 45.7 / 45.7 & 22.8 / 22.8 & 9.2 / 9.2 \\
Plan-and-Budget & 21.0 / 21.0 & 58.3 / 58.3 & 22.3 / 22.3 & 22.3 / 22.3 \\
\midrule
\multicolumn{5}{l}{\textit{A: Acquisition}} \\
IRCoT & 12.8 / 12.8 & 58.1 / 58.1 & 23.9 / 23.9 & 23.4 / 23.4 \\
Adaptive-RAG & 10.4 / 10.4 & 59.0 / 59.0 & 23.9 / 23.9 & 21.9 / 21.9 \\
R2AG & 19.7 / 19.7 & 50.8 / 50.8 & 23.5 / 23.5 & 27.7 / 27.7 \\
Q-RAG & 4.0 / 4.0 & 59.4 / 59.4 & 23.3 / 23.3 & 20.6 / 20.6 \\
\midrule
\multicolumn{5}{l}{\textit{U: Curation}} \\
LongLLMLingua & -- & 31.0 / 31.0 & 20.1 / 20.1 & -- \\
LLMLingua-2 & -- & 49.8 / 49.8 & 23.1 / 23.1 & -- \\
RECOMP & -- & 28.6 / 28.6 & 17.4 / 17.4 & -- \\
ACON & -- & 51.4 / 51.4 & 21.3 / 21.3 & -- \\
ACE & -- & 58.1 / 58.1 & 23.3 / 23.3 & -- \\
MCE text-only & -- & 58.2 / 58.2 & 23.6 / 23.6 & -- \\
SkillOpt & -- & 57.8 / 57.8 & 23.2 / 23.2 & -- \\
\bottomrule
\end{tabular}
\end{table}

Table~\ref{tab:auxiliary-four} shows that methods from the same family
can perform quite differently across tasks. Among the acquisition
methods, Q-RAG is one of the strongest on document question answering
but has the lowest score on BRIGHT. R2AG scores above \code{OPT0} on
both retrieval benchmarks but below it on document question answering.
Knowing that a task supports acquisition methods therefore tells us
which methods can run, but not which one to choose.
For methods adapted to produce search queries, the Reader control
provides a more direct comparison: it uses the same query interface
without input optimization. Plan-and-Solve and R2AG both score above
this control on both retrieval benchmarks. Plan-and-Budget scores
above it on one benchmark and below it on the other.

\section{When Does Conditional Routing Help?}
\label{app:routing-theory}
Different actions may work best on different tasks. Whether a router
can use this variation depends on what it knows about the task before
execution and how it uses that information to choose an action.

\begin{ailassumption}[Comparison and information]
\label{ass:ail-routing}
Fix an arena, backbone, profile $\theta$, and evaluation distribution.
The finite action set $\mathcal M$ contains \code{OPT0} and the five representatives,
all eligible on the comparison support. Let $U_m\in[0,1]$ be the Valid
utility for action $m$, using the task-group and paired reference
aggregation in Supplement~\ref{app:selector}. Each action retains its declared
downstream policy and resource profile. We condition on the fitted predictors
and frozen policy $\pi(Z)$, where $Z=\phi(x)$ contains only pre-action features.
\end{ailassumption}

Let $\mu_m(Z)=\mathbb E[U_m\mid Z]$. The best fixed action, the best
feature-based choice, and outcome-informed selection have values
\begin{equation}
\begin{aligned}
V_{\mathrm{fixed}}&=\max_{m\in\mathcal M}\mathbb E[U_m],\\
V_Z^\star&=\mathbb E[\max_{m\in\mathcal M}\mu_m(Z)],\\
V_{\mathrm{oracle}}&=\mathbb E[\max_{m\in\mathcal M}U_m].
\end{aligned}
\label{eq:ail-values}
\end{equation}
For example, suppose two task types occur equally often and have
identical visible features. One action scores one only on the first
type, another scores one only on the second, and all other scores
are zero. Choosing in hindsight gives $V_{\mathrm{oracle}}=1$.
Before execution, the router cannot distinguish the types, so
$V_Z^\star=V_{\mathrm{fixed}}=1/2$.

\begin{ailproposition}[Information and prediction]
Under Assumption~\ref{ass:ail-routing},
$V_{\mathrm{fixed}}\le V_Z^\star\le V_{\mathrm{oracle}}$.
For utility predictions $\widehat\mu_m(Z)$, define
\begin{equation}
\begin{aligned}
\varepsilon
  &=\mathbb E[\max_m|\widehat\mu_m(Z)-\mu_m(Z)|],\\
\eta_\pi
  &=\mathbb E[\max_m\widehat\mu_m(Z)-\widehat\mu_{\pi(Z)}(Z)].
\end{aligned}
\label{eq:ail-errors}
\end{equation}
Here $\varepsilon$ measures prediction error across actions, and
$\eta_\pi$ is the average gap between the highest predicted score
and the predicted score of the chosen action. If these expectations
are finite, then
\begin{equation}
V(\pi)-V_{\mathrm{fixed}}
\ge V_Z^\star-V_{\mathrm{fixed}}-2\varepsilon-\eta_\pi,
\label{eq:ail-bound}
\end{equation}
where $V(\pi)=\mathbb E[U_{\pi(Z)}]$. Thus, the router outperforms
the best fixed action whenever the gain available from pre-action
features, $V_Z^\star-V_{\mathrm{fixed}}$, exceeds
$2\varepsilon+\eta_\pi$.
\end{ailproposition}

\begin{proof}
For every fixed $a$, $\max_m\mu_m(Z)\ge\mu_a(Z)$. Also
$U_m\le\max_a U_a$, so
$\max_m\mu_m(Z)\le\mathbb E[\max_a U_a\mid Z]$.
Taking expectations proves the value ordering.

Write $e(Z)=\max_m|\widehat\mu_m(Z)-\mu_m(Z)|$. Comparing a
maximizer of $\mu_m(Z)$ with $\pi(Z)$ incurs at most two such errors,
giving
\[
\begin{aligned}
\max_m\mu_m(Z)-\mu_{\pi(Z)}(Z)
&\le 2e(Z)+\max_m\widehat\mu_m(Z)\\
&\quad-\widehat\mu_{\pi(Z)}(Z).
\end{aligned}
\]
Because $\pi$ uses only $Z$, $V(\pi)=\mathbb E[\mu_{\pi(Z)}(Z)]$.
Taking expectations and rearranging yields Equation~\ref{eq:ail-bound}.
\end{proof}

\noindent\textbf{Experimental connection.}
Fig.~\ref{fig:motivation}(b) asks how much could be gained by
choosing the best observed action for each task rather than using
one fixed action selected on development data.
Fig.~\ref{fig:selector}(a) asks a different question: does choosing
from pre-action features outperform random routing at the same
intervention frequency? The bound compares against the
population-best fixed action. It explains why a large gain in
hindsight need not translate into a large gain for a fitted router.
Fig.~\ref{fig:selector}(b) tests which groups of pre-action features
help the fitted router choose. Even with informative features,
prediction error can reduce the gain. A threshold, margin, or
Guardrail rule can also select an action other than the one with
the highest predicted score; $\eta_\pi$ is zero when the router
always selects a predicted maximizer. The same argument applies
to predictors of gain over \code{OPT0}, since subtracting the
same reference score $U_0$ from every action leaves their
ordering unchanged.
\end{document}